\documentclass[11pt]{article}
\usepackage[margin=1in]{geometry}
\usepackage{amsfonts}
\usepackage{amsmath,amsthm,amssymb}
\usepackage{latexsym}
\usepackage{epic}
\usepackage{epsfig}
\usepackage{hyperref}
\usepackage{verbatim}
\usepackage[justification=centering]{caption}
\usepackage{enumitem}
\usepackage{color}
\allowdisplaybreaks[1]
\usepackage[numbers]{natbib}
\usepackage{algorithm}
\usepackage{algorithmic}
\usepackage{setspace}

\usepackage{tikz}
\usepackage{graphicx}
\usepackage{tikzscale}
\usepackage{wrapfig}
\usepackage{float}
\usetikzlibrary{arrows.meta,arrows}
\usetikzlibrary{decorations.pathreplacing}
\usetikzlibrary{positioning,chains,fit,shapes,calc}
\usepackage{wrapfig}
\usepackage{pgfplots}
\pgfplotsset{compat = newest}

\usepackage{caption}
\usepackage{subcaption}

\usepackage{xspace}

\newcommand{\cross}{\times}

\newcommand{\set}[1]{\left\{ #1 \right\}}

\renewcommand{\tilde}{\widetilde}

\newcommand{\bvec}[1]{\boldsymbol{ #1 }}

\def\min{\qopname\relax n{min}}

\def\Ex{\qopname\relax n{\mathbf{E}}}

\newcommand{\RR}{\mathbb{R}}

\def\C{\mathcal{C}}
\def\D{\mathcal{D}}

\def\H{\mathcal{H}}

\def\X{\mathcal{X}}
\def\Y{\mathcal{Y}}
\def\Z{\mathcal{Z}}

\def\sse{\subseteq}

\newcommand{\eat}[1]{}

\newenvironment{lp*}{\begin{equation*}  \begin{array}{lll}}{\end{array}\end{equation*}}

\newcommand{\tst}{\mathtt{test}}
\newcommand{\ag}{\mathtt{Ag}}
\newcommand{\bp}{\mathtt{Bp}}

\newcommand{\y}{\bvec{y}}

\newcommand{\dist}{\mathtt{dist}}

\title{PAC Learning is just Bipartite Matching\\ (Sort of) 
}

\author{
  Shaddin Dughmi\thanks{Supported by NSF Grant CCF-2009060} \\
Department of Computer Science\\
University of Southern California\\
{\tt shaddin@usc.edu}
}

\begin{document}

\maketitle


\section{Introduction}
The main goal of this article is to convince you, the reader, that supervised learning in the Probably Approximately Correct (PAC) model is closely related to --- of all things --- bipartite matching! 
\mbox{En-route} from PAC learning to bipartite matching, I will overview a particular \emph{transductive model} of learning, and associated \emph{one-inclusion graphs}, which can be viewed as a generalization of some of the \emph{hat puzzles} that are popular in recreational mathematics. Whereas this transductive model is far from new, it has recently seen a resurgence of interest as a tool for tackling deep questions in learning theory. In hindsight, a secondary purpose of this article could be as a (biased) tutorial on the connections between the PAC and transductive models of learning.

I found these perspectives on learning surprising and relatable, as an outsider who primarily thinks about discrete algorithms and combinatorial optimization.
I hope this article  can help others of similar disposition approach learning theory, and perhaps even help researchers uncover more connections between machine learning and classical questions in combinatorics and optimization.

In what sense is PAC learning ``just matching''? For classification problems, quite literally: Multiclass classification in the PAC model is approximated by an implicitly-described bipartite matching problem, albeit a huge one. 
Mathematicians and computer scientists know a lot about the structure of bipartite matchings, and I will demonstrate how we can invoke this understanding to deduce properties of learning problems and algorithms.

Outside of classification, e.g. for regression or problems with structured multivariate labels, the connection to matching is less direct. The learning task is approximated by an implicitly-described optimization problem which resembles a (non-linear) fractional generalization of bipartite matching. While this seems to preclude applying ideas from matching theory directly, it suggests that starting from what you know about matching and generalizing it appropriately might uncover useful structure. I will outline one example of this approach, and speculate wildly  about others.

\subsection*{Outline}
This article is structured as follows. First, I will recap the basics of supervised learning and the PAC model. Then I strip away many of the details of learning by reducing to a particular \emph{transductive model}. This is a bare-bones, distribution-free model of supervised learning which has its roots in the earliest days of the field \cite{vapnik_theory_1974,vapnik_estimation_1982,haussler_predicting_1994}. The model also bears striking similarities, but also important differences, with \emph{hat puzzles} from recreational mathematics.  I will try to convince you that this transductive model  is ``close enough,'' for most intents and purposes, to the PAC model.

Once in the transductive model, we will get into the most accessible of supervised learning problems: classification. I will review the \emph{one-inclusion graph} \cite{bondy_induced_1972,alon_partitioning_1987,haussler_predicting_1994}, which exactly encodes transductive classification as a combinatorial optimization problem. I will reinterpret this optimization problem as bipartite matching by a simple change of perspective \cite{asilis_regularization_2024}, and describe some implications of this matching perspective for understanding the space of near-optimal classification algorithms, as well as characterizing the optimal learning rate. 

We will then move on to more general supervised learning problems, to include things like regression or pretty much anything else. I will present a generalization of the one-inclusion graph that again exactly encodes transductive learning, and argue that this generalization is still intimately tied to bipartite matching. To drive that point home, I will present one example of how we start from a deep fact about bipartite matching --- its \emph{compactness} in a set-theoretic sense --- and generalize it appropriately to derive compactness for the sample complexity of learning~\cite{asilis_transductive_2024}.

Finally, I will close with irresponsible speculation about future applications of this perspective.

\subsection*{Disclaimer} This article explores one perspective on PAC learning which I personally found helpful,  and did not see written down in one place as I attempt to do here. More standard perspectives and techniques include dimensional characterizations of learnability, sample compression, uniform convergence, and others. Discussion of those at any length is beyond the scope of this article, and above my pay grade.  For a well rounded treatment of machine learning theory, by actual experts, refer to any of the excellent texts on the topic such as \citet{shalev-shwartz_understanding_2014}, \citet{mohri_foundations_2018}, and \citet{anthony_neural_1999}. 


\section{Basic Background: Supervised Learning and the PAC Model}
\label{sec:background}

At this point almost everyone has heard of machine learning (ML). Anyone likely to stumble upon this article will have also heard of its most influential special case, supervised learning, and those theoretically inclined will also be familiar with the PAC model. Nonetheless, I will set the stage by  recapping the basics.

\subsection{Basics of Supervised Learning}

\emph{Supervised Learning} is the task of ``coming up'' with a function $f: \X \to \Y$ to ``explain'' or ``fit'' a sequence of input/output examples   $(x_1,y_1), \ldots, (x_n,y_n)$, with $x_i \in \X$ and $y_i \in \Y$.  Here $\X$ is a \emph{data domain} consisting of \emph{datapoints} $x \in \X$, $\Y$ is a \emph{label set} consisting of \emph{labels} $y \in \Y$, and the sequence $(x_1,y_1),\ldots,(x_n,y_n)$ is the \emph{training data} consisting of \emph{labeled examples (a.k.a.~samples)}~$(x_i,y_i)$.  I~will refer to the chosen function $f$ as a \emph{predictor}, and to $n$ as the \emph{sample size}. A \emph{learning algorithm} takes as input training data, and outputs (some representation of) a predictor $f \in \Y^\X$.\footnote{Note that this describes the usual \emph{batch}, a.k.a.~\emph{offline}, setting of supervised learning. I do not discuss other paradigms such as online or active learning in this article.}

Success in supervised learning is defined as \emph{generalization} to  future examples: For a typical \emph{test example}  $(x_{\tst},y_{\tst})$, the predicted label $y'_{\tst}=f(x_{\tst})$ should ``equal'' $y_{\tst}$, perhaps approximately. We usually assume the test example is drawn from the same  ``source'' as the training data  --- commonly, i.i.d.~from the same distribution. The quality of the prediction is quantified by $\ell(y'_{\tst},y_{\tst})$, where $\ell:~\Y~\times~\Y \to \RR_{\geq 0}$ is a \emph{loss function} chosen as part of the problem definition. Common loss functions include the 0-1 loss $\ell_{0-1}(y',y) = [y' \neq y]$ for \emph{classification} problems,\footnote{The notation $[P]$ denotes $1$ when predicate $P$ is true, and denotes $0$ when $P$ is false.} as well as the absolute loss $|y'-y|$ or squared loss $(y'-y)^2$ for \emph{regression problems} featuring $\Y  \sse \RR$.

Nontrivial generalization properties are typically only possible if one assumes something about the data.\footnote{The need for such an assumption is formalized by the  \emph{no free lunch theorems} of supervised learning \cite{wolpert_connection_1992,wolpert_lack_1996,schaffer_conservation_1994}.} The Bayesian approach to  machine learning, common in many applications, assumes some parametric form for the distribution generating the data, and postulates a prior on the parameters. This is not the approach I will take in this article. Instead, I will focus on the frequentist --- and some would say ``worst-case'' or ``adversarial'' ---  approach that is common in the computational learning theory community, embodied by the PAC model. Here we assume that the (training and test) data can be explained, perhaps approximately, by a function in some ``simple enough to learn'' class of functions $\H \sse \Y^\X$, often called the \emph{hypotheses}. Equivalently, we  seek a predictor which explains the unseen data roughly  as well as the best hypothesis $h^* \in \H$, whether or not we assume that $h^*$ itself provides a perfect explanation.

 \paragraph{Common Algorithmic Templates.} Perhaps the best known general-purpose supervised learning algorithm is \emph{empirical risk minimization (ERM)}, which chooses as its predictor a hypothesis $f \in \H$ minimizing $\frac{1}{n} \sum_{i=1}^n \ell(f(x_i),y_i)$ --- a quantity called the \emph{training error}, \emph{empirical error}, or \emph{empirical risk} of $f$. 
A common template for generalizing ERM involves adding a \emph{regularization term} $\psi(f)$ to the  objective function, typically chosen to measure some notion of ``hypothesis complexity.'' An algorithm instantiating this template is known as a \emph{structural risk minimizer (SRM)}, and chooses as its predictor the hypothesis $f \in \H$ minimizing the \emph{structural risk} $\frac{1}{n} \sum_{i=1}^n \ell(f(x_i),y_i) + \psi(f)$. Other well-known algorithms, such as gradient descent and its variations,  can frequently be interpreted as approximate implementations of ERM or SRM.

\paragraph{Proper vs Improper Learning.} A learning algorithm is said to be \emph{proper} if its predictor $f$ is always chosen from the hypothesis class, i.e., $f \in \H$, otherwise it is said to be \emph{improper}. ERM  is an example of a proper learning algorithm, as are SRM algorithms of the form described above.  In the \emph{proper regime} of learning, algorithms are required to be proper. This article will be concerned with the more flexible \emph{improper regime} (a.k.a.~\emph{representation-independent learning}), where no such constraint is placed on the learner. In other words, all we care about is predictive power at test time, rather than any insights derived from the functional form or representation of the predictor~itself.

\subsection{The PAC Model}
A standard mathematical setup for evaluation of supervised learning algorithms, at least in the theoretical computer science community, is Valiant's \emph{Probably Approximately Correct (PAC) model} of learning (see e.g.~\cite{kearns_introduction_1994,mohri_foundations_2018}). Here, we assume there is an unknown distribution $\D$ on $\X \times \Y$ from which training and test data are  drawn.  Specifically, the labeled datapoints of the training set  $(x_1,y_1), \ldots, (x_n,y_n)$, as well as the test data  $(x_\tst,y_\tst)$, are i.i.d.~from $\D$. Often it is assumed that $\D$ lies in some class of distributions of interest. The \emph{true expected loss}, or simply \emph{loss}, of a predictor $f: \X \to \Y$ is the expected loss it incurs on draws from $\D$, written $L_\D(f) = \Ex_{(x,y) \sim \D} \ell(f(x),y)$.

There are two main ``settings'' in PAC learning. The  \emph{realizable setting} only requires that the data be perfectly explained by some hypothesis in $\H$. More generally, the \emph{agnostic setting} makes no assumption relating the data to the hypotheses, but shifts the goalposts as necessary to allow nontrivial guarantees: the expected loss at test time is evaluated only ``relative'' to that of the best hypothesis $h^* \in \H$. There are other settings which make more nuanced assumptions, such as $\D$ being of a particular parametric form or its support living in some (unknown) lower-dimensional space, etc. I will mostly discuss the realizable and agnostic settings in this article, those being the simplest and most studied from a theoretical perspective. 

The PAC model demands high probability guarantees of learners, in the worst case over distributions of interest. Consider first the realizable setting, where $\D$ is such that $\min_{h \in \H} L_{\D}(h) = 0$. A PAC learner has \emph{error} $\epsilon=\epsilon(n)$ and \emph{confidence} $\delta=\delta(n)$ if, when training data consists of $n$ i.i.d~samples from a realizable distribution $\D$, it produces a predictor $f$  satisfying $L_\D(f) \leq \epsilon$ with probability at least $1-\delta$. In the agnostic setting, where $\D$ can be arbitrary, we require $L_\D(f) - \min_{h \in \H} L_\D(h) \leq \epsilon$ with probability $1-\delta$.

In both the realizable and agnostic settings, we look for PAC learners with small $\epsilon$ and $\delta$ as a function of the sample size $n$. An equivalent perspective looks at the sample complexity $m(\epsilon,\delta)$, which is the minimum sample size which guarantees error  at most $\epsilon$ with probability at least $1-\delta$. We say a problem is \emph{PAC learnable} if its PAC sample complexity is finite whenever $\epsilon,\delta > 0$.

For most PAC learning problems, learnability and sample complexity are characterized in terms of a  ``dimension'' of the hypothesis class. Most prominently this is the \emph{VC dimension} for binary classification, the \emph{fat shattering dimension} for agnostic regression, and the \emph{DS dimension} for multiclass classification (see \cite{anthony_neural_1999,daniely_optimal_2014,brukhim_characterization_2022}). Treatment of these is beyond the scope of this article. The unfamiliar reader need not worry, however,  as dimensions will feature only tangentially in our~discussion.



\section{A Transductive Model of Learning, and why it's Good Enough}
\label{sec:trans}

I will now present the learning model that will be our main playground in this article, then try to convince you that we lose little by restricting our attention to it. This model was first employed by \citet{haussler_predicting_1994}, and has been quite busy in recent years, with its learners used as precursors to PAC learners (e.g. \cite{daniely_optimal_2014,brukhim_characterization_2022,aden-ali_optimal_2023,daskalakis_is_2024,asilis_regularization_2024,montasser_adversarially_2022}).
I will describe the model as conceptualized by \citet{daniely_optimal_2014} and further developed by  \citet{asilis_regularization_2024}, and refer to it simply as \emph{the transductive model}  in keeping with some of the recent literature.   
%
Truth be told, however,  the transductive approach to learning --- first explicitly articulated by \citet[Chapter 6.1]{vapnik_nature_1998} --- is much broader than just this particular model. Most notably, the transductive model I use in this article is concerned with making only a single prediction, and presumes data is chosen by a particular strong adversary. We will come back to transduction more generally later.   

\subsection{The Transductive Model}
The (single-prediction, adversarial) transductive model is  a game of ``fill in the blank'' played between a learner and an adversary. For a given positive integer $n$, the game proceeds as follows:
\begin{enumerate}
\item The adversary chooses a \emph{labeled dataset} $(x_1,y_1), \ldots,(x_n,y_n)$, with $x_i \in \X$ and $y_i \in \Y$. \\(The adversary may  be constrained in their choice of dataset --- more on this shortly.)
\item One label $y_i$, chosen uniformly at random, is hidden. The remaining data \[(x_1,y_1),\ldots,(x_{i-1},y_{i-1}),(x_i,?),(x_{i+1},y_{i+1}),\ldots,(x_n,y_n),\] where ``?'' is a ``blank'' symbol obscuring $y_i$, is displayed to the learner. 
\item The learner is prompted to make a prediction $y'_i$ for the label of $x_i$, i.e., to ``fill in the blank'', incurring loss $\ell(y'_i,y_i)$.
\end{enumerate}

The adversary in the transductive model chooses the training and test data, but cannot control which is which: the identity of the test example $(x_i,y_i)$ is chosen uniformly at random. Then, the learner is fed training data   $\set{(x_j,y_j) : j \neq i}$ and prompted with the test datapoint $x_i$. In relation to  PAC learning, where training and test data are sourced i.i.d.~from some distribution, the transductive model can be viewed as ``hard-coding'' a uniform distribution over $n$ examples then drawing training and test data \emph{without replacement}.

We will distinguish two settings of the transductive model, \emph{realizable} and \emph{agnostic}, analogous to the same in the PAC model. The transductive model was originally considered in the realizable setting~\cite{haussler_predicting_1994,daniely_optimal_2014}, where the adversary is constrained to instances $(x_1,y_1), \ldots, (x_n,y_n)$ which are perfectly explained by a some hypothesis $h \in \H$, meaning $h(x_j) = y_j$ for all examples. The (realizable) \emph{transductive error}  of a learner on datasets of size $n$ is simply the expected loss it incurs in this game, i.e., for a worst-case realizable instance $(x_1,y_1), \ldots, (x_n,y_n)$ chosen by the adversary. Here, the expectation is over the random identity of the test example $(x_i,y_i)$ as well as any internal randomness to the learner. In the agnostic setting of this model, as articulated in~\cite{asilis_regularization_2024}, the adversary is completely unconstrained in their choice of instance $(x_1,y_1), \ldots, (x_n,y_n)$. We then shift the goalposts by subtracting the best-in-class loss on the same instance, as necessary to allow for nontrivial  learning: the \emph{agnostic transductive error} is the expected loss incurred by the learner minus $\min_{h \in \H} \frac{1}{n} \sum_{i=1}^n \ell(h(x_i),y_i)$.

Whether in the realizable or agnostic setting, we look for learners with small transductive error rate $\epsilon(n)$, as a function of the dataset size $n$. An equivalent perspective looks at the sample complexity $m(\epsilon)$, which is the minimum dataset size guaranteeing  error at most $\epsilon$. We say a problem is \emph{learnable} in the transductive model if its transductive sample complexity is finite for every $\epsilon > 0$.

\begin{figure}
  \centering
    \begin{subfigure}{0.3\textwidth}
        \centering
        \scalebox{0.8}{\begin{tikzpicture}
    \coordinate (A) at (1,0.5);
    \coordinate (B) at (-1,0);
    \coordinate (C) at (0,1);
    \coordinate (D) at (0.3,-1);

    \node[red] at (A) {\textbf{+}};
    \node[red] at (B) {\textbf{+}};
    \node[red] at (C) {\textbf{+}};
    \node[red] at (D) {\textbf{+}};

    \coordinate (rectSW) at (-1.5,-1.5); 
    \coordinate (rectNE) at (1.5,1.5);   

   \draw[rectangle, draw] (rectSW) rectangle (rectNE);

    \foreach \i in { 20,30,40,50,60,120,130,140,150,160,-20,-30,-40,-50,-60,-120,-130,-140,-150,-160} {
        \pgfmathsetmacro\xcoord{3*cos(\i)}
        \pgfmathsetmacro\ycoord{3*sin(\i)}
        \node[blue] at ({\xcoord},{\ycoord}) {\textbf{-}};
    }
\end{tikzpicture}

}
        \caption{Labeled dataset consistent with an axis-aligned rectangle.}
        \label{fig:rectangles_a}
    \end{subfigure}
   \hfill 
    \begin{subfigure}{0.3\textwidth}
        \centering
        \scalebox{0.8}{\begin{tikzpicture}
    \coordinate (A) at (1,0.5);
    \coordinate (B) at (-1,0);
    \coordinate (C) at (0,1);
    \coordinate (D) at (0.3,-1);

    \node[red] at (A) {\textbf{+}};
    \node[red] at (B) {\textbf{+}};
    \node[red] at (C) {\textbf{+}};
    \node[red] at (D) {\textbf{+}};

    \coordinate (rectSW) at (-1.5,-1.5); 
    \coordinate (rectNE) at (1.5,1.5);   

   \draw[rectangle, draw] (rectSW) rectangle (rectNE);

    \foreach \i in { 20,30,40,60,120,130,140,150,160,-20,-30,-40,-50,-60,-120,-130,-140,-150,-160} {
        \pgfmathsetmacro\xcoord{3*cos(\i)}
        \pgfmathsetmacro\ycoord{3*sin(\i)}
        \node[blue] at ({\xcoord},{\ycoord}) {\textbf{-}};
    }

        \pgfmathsetmacro\xcoord{3*cos(50)}
        \pgfmathsetmacro\ycoord{3*sin(50)}
        \node[black] at ({\xcoord},{\ycoord}) {\textbf{?}};

        \draw[rectangle, draw] (-1.4,-1.4) rectangle (\xcoord+0.16,\ycoord+0.18);

\end{tikzpicture}

}
        \caption{When any label is omitted, both + and - are consistent with some rectangle.}
        \label{fig:rectangles_b}
      \end{subfigure}
      \hfill
    \begin{subfigure}{0.3\textwidth}
      \centering
      \scalebox{0.8}{\begin{tikzpicture}
    \coordinate (A) at (1,0.5);
    \coordinate (B) at (-1,0);
    \coordinate (C) at (0,1);
    \coordinate (D) at (0.3,-1);
        \coordinate (E) at (0,0);

    \node[red] at (A) {\textbf{+}};
    \node[red] at (B) {\textbf{+}};
    \node[red] at (C) {\textbf{+}};
    \node[red] at (D) {\textbf{+}};
    \node[red] at (E) {\textbf{+}};

    \coordinate (rectSW) at (-1.05,-1.05); 
    \coordinate (rectNE) at (1.05,1.05);   

   \draw[rectangle, draw] (rectSW) rectangle (rectNE);

    \foreach \i in { 20,30,40,50,60,120,130,140,150,160,-20,-30,-40,-50,-60,-120,-130,-140,-150,-160} {
        \pgfmathsetmacro\xcoord{3*cos(\i)}
        \pgfmathsetmacro\ycoord{3*sin(\i)}
        \node[blue] at ({\xcoord},{\ycoord}) {\textbf{-}};
    }

\end{tikzpicture}

}
      \caption{The minimal axis-aligned rectangle is sensitive only to the four + points defining it.}
      \label{fig:rectangles_c}
    \end{subfigure}
    \caption{Realizable binary classification for axis aligned rectangles in the transductive model.}
    \label{fig:rectangles}
\end{figure}
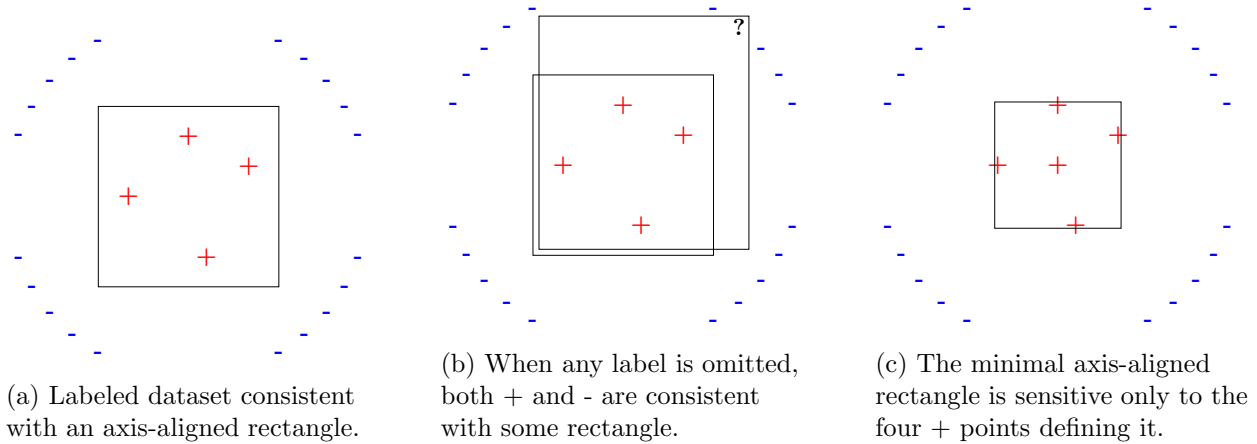

\paragraph{Example: Axis-aligned Rectangles.} To illustrate the transductive model, consider realizable binary classification of axis-aligned rectangles as described in~\cite{haussler_predicting_1994} and illustrated in Figure~\ref{fig:rectangles}. The adversary selects $n$ datapoints in Euclidean space $\RR^d$, and labels them  in a manner consistent with some axis-aligned rectangle as in Figure~\ref{fig:rectangles_a}. A random label is then omitted, and the learner is prompted to predict this label. A learner which merely minimizes empirical risk --- i.e., arbitrarily chooses an axis-aligned rectangle which is consistent with observed labels, and fills in the missing label accordingly --- can make a mistake with probability $1$. Such a scenario is illustrated in Figure~\ref{fig:rectangles_b}. This stands in contrast to the PAC model, where empirical risk minimization guarantees a misclassification rate of $O\left(\frac{d}{n}\right)$ with high probability, since axis-aligned rectangles in $\RR^d$ have a VC dimension of $2d$. A more careful choice of learning rule guarantees an error probability of $\frac{2d}{n}$ also in the transductive model: choose the \emph{smallest} axis-aligned rectangle consistent with the observed labels, and fill in the missing label accordingly. This is illustrated in Figure~\ref{fig:rectangles_c}.

\paragraph{Analogy to Hat Puzzles.} The reader familiar with \emph{hat puzzles} (see e.g. \cite{krzywkowski_hat_2010,butler_hat_2009}) might notice a similarity to the transductive model. There are many variations of hat puzzles, the most relevant to us of which look something like this: There are $n$ players and a supply of hats of $k$ different colors. An adversary places a hat on the head of each of the players. Each player must then guess the color of their own hat by looking at all, or in some formulations some, of the hats of the other players. The players formulate a joint strategy in advance, and a typical goal is to maximize the number of correct guesses.

When the transductive model is applied to classification in the realizable setting, the  analogy to hat puzzles is clear: the players are $x_1,\ldots,x_n$, the hat colors are $y_1,\ldots,y_n$, and player $i$'s guessing task corresponds to the scenario in which $y_i$ is hidden. The learner  therefore corresponds to the joint strategy of the players, and transductive error is proportional to the number of players guessing incorrectly. 

There are important differences between hat puzzles and the transductive model, however. To my knowledge, most hat puzzles permit the adversary to assign hat colors either arbitrarily or uniformly at random. The transductive model, in contrast, is constrained to the hat assignments permitted by the hypothesis class (in the realizable setting, at least). Another important difference is that hat puzzles are commonly concerned with deterministic guessing strategies.  This is because the optimal randomized strategy against such a powerful adversary is the trivial one: each player randomly guesses their hat color. Obtaining a similar guarantee with a deterministic strategy is therefore the more interesting puzzle here. 

To my knowledge, the analogy between hat puzzles and transductive learning appears completely absent from the literature. Despite their differences, there might yet be fruitful connections between the two areas. One tantalizing tidbit is that the hypercube  perspective on hat puzzles articulated by \citet{butler_hat_2009} appears to be a special case of the \emph{one-inclusion graphs} that are almost synonymous with the transductive model --- more on these graphs in Section~\ref{sec:class}.

\paragraph{A Note on Transduction vs Induction, more generally.} The usual  conception of supervised learning is as a problem of \emph{inductive inference}: reasoning from particular observations (the training data) to a general rule (the chosen predictor), which can then be applied to answer specific questions at test time (labeling the test data). A (seemingly) more permissive approach is to care not for the rule itself, but for its evaluation at the particular points of interest; i.e.,  we reason from particular observations (the training data) to answer specific questions (labeling the test data). This is \emph{transductive learning} as first explicitly articulated by~\citet{vapnik_nature_1998}, though the essential ideas date back to \citet{vapnik_theory_1974} and \citet{vapnik_estimation_1982}. Notice that this general transductive approach does not presuppose a particular data model or objective, and can be considered in other setups such as the PAC model, online learning, etc.

When there is only a single test point, and learners are allowed to be improper, induction and transduction are essentially equivalent. Indeed,  the formal distinction between the two becomes simply a matter of \emph{currying}\footnote{Currying is a notion common in functional programming, and amounts to the observation that a function \linebreak $g: A \times B \to \C$ can equivalently be viewed as a function $g': A \to (B \to C)$, with $g'(a)$ itself a function from $B$ to $C$.}  --- see the related discussion in~\cite{montasser_transductive_2022}. This equivalence no longer holds in general when multiple test datapoints must be labeled, nor in the proper regime. Since we operate in the improper regime with a single test example, the distinction between transduction and induction becomes merely a matter of perspective for our purposes. Nonetheless, the  transductive model is best thought of as reasoning for a specific test example, hence the name.

A related perspective on the distinction between transduction and induction  is the following. By catering the learned predictor to a specific test example, transductive learners are ``improper by default'' when viewed across all possible test points. Indeed, by focusing only on prediction and freeing the learner completely from representing any general rule, the transductive perspective unleashes the full power of improper learning. Induction, in contrast, is most meaningful when one favors rules of a particular form, as is the case in the proper regime.

\subsection{Relationship to the PAC Model}
\label{subsec:pac_trans}
Despite seeming  incomparable at first glance, the PAC and transductive models turn out to be closely related.\footnote{This is specific to the  improper (i.e., unrestricted) regime of PAC learning, which is our focus in this article. The transductive model, in allowing improperness by default, is more permissive than proper PAC learning~\cite{daniely_optimal_2014}.} Speaking qualitatively, learnability is equivalent in the two models, and this holds for both realizable and agnostic learning problems with bounded loss functions.

Quantitatively, we can say something quite strong in the realizable setting with bounded losses: The two models are equivalent up to  low-order factors in the error and the sample complexity. This follows from  efficient black-box reductions  between the two models, one in each direction. We therefore lose very little, and gain quite a lot through the connection to matching and otherwise, by studying the transductive model instead of the PAC model for realizable learning problems.

The exact quantitative relationship between the two models  is more unsettled in the agnostic setting: Whereas it appears that PAC learning is essentially  no harder than  transductive learning (up to low-order terms, for most natural loss functions, through an efficient black-box reduction), it remains plausible that transductive sample complexity exceeds its PAC counterpart by a multiplicative factor on the order of $\frac{1}{\epsilon}$, even for classification problems.
This potential gap of $\frac{1}{\epsilon}$ in agnostic sample complexities is the main asterisk to the stance I take in this article of treating the PAC and transductive models as interchangeable. Given  the transductive model's  fruitful connections to basic graph-theoretic questions, as well as the possibility that future work could eliminate this gap, I hope the reader finds this qualification forgivable.

I will now recap both the realizable and agnostic states of affairs in more detail.

\subsubsection*{The Realizable Setting}
  In the realizable setting with losses in $[0,1]$, the transductive and PAC models are equivalent up to  low-order factors in the error and the sample complexity. This is somewhat folklore knowledge in the field, but the proofs are written down explicitly in~\cite{asilis_regularization_2024}.  I'll recap those proofs informally next.

For one direction of the equivalence, a realizable PAC learner guaranteeing error $\epsilon$ with probability $1-\delta$ on $n$ samples can be converted to a realizable transductive learner guaranteeing error $O(\epsilon + \delta)$ on datasets consisting of $n$ samples. 
This is done in pretty much the obvious way: Given a transductive query of the form $(x_1,y_1), \ldots,(x_i,?), \ldots,(x_n,y_n)$, we train a PAC learner with $n$ i.i.d.~uniform draws from $\set{(x_j,y_j)}_{j \neq i}$, then invoke the resulting predictor on the test point $x_i$. For the error guarantee, observe two key facts: First, the PAC learner has expected loss at most $\epsilon + \delta$ when its $n$ training samples as well as its test point are i.i.d. uniform draws from $\set{(x_j,y_j)}$. Second, i.i.d. uniform draws from $\set{(x_j,y_j)}$ happen to exclude the transductive test point $(x_i,y_i)$ with probability $\frac{1}{e}$. Thus, restricting the PAC learner to training sets excluding $x_i$ --- as we do in this reduction --- multiplies the loss bound of $\epsilon + \delta$  by a factor of at most $e$, by Markov's inequality.

For the other direction, a realizable transductive learner guaranteeing error $\epsilon$ on datasets of size $n$ can be  converted to a realizable PAC learner which guarantees error $O(\epsilon)$ with probability $1-\delta$ when given $O(n \log \frac{1}{\delta})$ samples. This again is fairly elementary: The transductive learner guarantees expected loss $\epsilon$ on a PAC instance via a \emph{leave one out} argument --- essentially conditioning on the training and test data and invoking the principle of deferred decisions with regards to which is which. This can then be boosted to a high probability guarantee, as required in PAC learning, by repetition $O(\log \frac{1}{\delta})$ times. Markov's inequality and the union bound show that one of the predictors has loss $O(\epsilon)$ with probability $1-\frac{\delta}{2}$, and a small hold-out set of examples can be used to identify such a predictor with probability $1-\frac{\delta}{2}$.

For the large class of metric loss functions, a more sophisticated  reduction gets rid of the multiplicative blowup in sample complexity \cite{aden-ali_optimal_2023,dughmi_is_2024}. Specifically, the number of samples is reduced from $O\left(n \log \frac{1}{\delta}\right)$ to $n+O\left(\frac{1}{\epsilon} \log \frac{1}{\delta}\right)$. This is possible through a more economical repetition construction which reuses examples across multiple invocations of the transductive learner.

\subsection*{The Agnostic Setting} 

In relating the transductive and PAC models in the agnostic setting, the best we can currently say for general bounded loss functions is the following:  Adaptations of the previously-described realizable reductions result in a degraded bound on the number of samples, by a factor of at most $O\left(\frac{1}{\epsilon}\right)$. To see why this might be, note that both those reductions rely on applying Markov's inequality to the (evidently nonnegative) loss of a randomized predictor: if its expected loss is $\epsilon$, then for an absolute constant $c \geq 1$ the realized predictor's loss exceeds $c \cdot \epsilon$ with probability at most $\frac{1}{c}$. Agnostic error discounts the loss of a predictor by the best-in-class loss, leading to a random variable which is no longer nonnegative, and therefore  no longer enjoys the same Markov guarantee. Indeed, when $\epsilon$ denotes the expectation of this discounted loss, Markov's inequality --- adapted appropriately to what is now a random variable in $[-1,1]$ --- guarantees that the discounted loss of the realized predictor exceeds $c \epsilon$ with probability at most $\frac{1+\epsilon}{1+ c \epsilon} \approx 1- (c-1) \epsilon = 1-O(\epsilon)$. In other words, the predictor is guaranteed to be ``good'' --- in the sense of having loss at most $c \epsilon$ --- with probability only $(c-1) \epsilon = O(\epsilon)$. 

Now consider what happens to the two reductions between the transductive and PAC models in their agnostic settings, in light of this degraded Markov guarantee. When converting a transductive learner to a PAC learner, we now need on the order of $\frac{1}{\epsilon} \log \frac{1}{\delta}$ repetitions --- in light of ``good'' predictors now arising with probability $O(\epsilon)$ --- in order to boost the expected error guarantee to a high probability one. For the other direction, when converting a PAC learner to a transductive learner using $n$ samples for each, Markov's inequality now fails to provides any nontrivial guarantees: the discounted loss of a PAC learner --- now a random variable in $[-1,1]$ --- can attain its maximum value of $1$ in the $\frac{1}{e}$-probability event that the test point is excluded, despite having small expectation~$\epsilon$. One remedy for this is to ``blow up'' the number of transductive samples to $m=\frac{n}{\epsilon}$, while still training the PAC learner with $n$ draws. In more detail, for a transductive query of the form $(x_1,y_1), \ldots,(x_i,?), \ldots,(x_m,y_m)$, we train a PAC learner with $n = \epsilon m$ i.i.d.~uniform draws from $\set{(x_j,y_j)}_{j \neq i}$, then invoke the resulting predictor on the test point $x_i$. Notice that $n$ draws from the uniform distribution on $\set{(x_j,y_j)}_{j=1}^m$ will exclude the test point $(x_i,y_i)$ with probability on the order of $1-\epsilon$. Using Markov's inequality in similar fashion, adapted appropriately to random variables in $[-1,1]$, the bound of $\epsilon + \delta$ on the discounted loss increases only to $O(\epsilon) + \delta$ when we condition on the event of interest: that where we exclude $x_i$ from the training data.  

To recap, we cannot yet rule out a multiplicative gap of $\frac{1}{\epsilon}$ between transductive and PAC sample complexities in general, in either direction. That said, this gap seems somewhat more fundamental in one of the directions than the other.  For the large class of metric loss functions, a transductive learner guaranteeing agnostic error $\epsilon$ on datasets of size $n$ can be converted to a PAC learner guaranteeing agnostic error $O(\epsilon)$ with probability $1-\delta$ when given $n+O\left(\frac{1}{\epsilon^2} \log \frac{1}{\epsilon\delta}\right)$ samples. This is through an adaptation of the economical repetition construction of \citet{aden-ali_optimal_2023} to the agnostic setting by  \citet{dughmi_is_2024}.
This suggests that PAC learning is essentially no harder than transductive learning for most natural loss functions. Whether the converse is true remains very much in question,  as discussed and explored in \cite{dughmi_is_2024}. Specifically, it remains plausible that transductive sample complexity exceeds PAC sample complexity by a multiplicative factor on the order of $\frac{1}{\epsilon}$, even for multiclass classification. Such a gap in sample complexity was ruled out for binary classification through a non-reduction approach in~\cite{dughmi_is_2024}, where we also conjecture that the gap can be  eliminated more generally.



\section{Classification, One-Inclusion Graphs, and Bipartite Matching}
\label{sec:class}

A \emph{classification problem} is a supervised learning problem with the 0-1 loss function $\ell_{0-1}(y',y)=[y' \neq y]$, which evaluates to $1$ when $y' \neq y$ and to $0$ when $y'=y$. It is important to note that this allows for an arbitrary domain $\X$, an arbitrary label set $\Y$, and an arbitrary hypothesis class $\H \sse \Y^\X$.  Classification therefore forms a rich and expressive class of problems,  garnering the lion's share of theoretical work on supervised learning (particularly in the PAC model), and featuring in essentially every application domain of machine learning. 
When there are only two labels, canonically $\Y= \set{0,1}$, we have a \emph{binary classification} problem.  More generally, \emph{multiclass classification} problems can have any number of labels, even infinitely many of any cardinality.

I will discuss classification problems in the transductive model, emboldened by its close relationship to the PAC model as described in Section~\ref{sec:trans}. I will start with a primer on one-inclusion graphs (OIGs), which encode transductive classification exactly  as a combinatorial optimization problem. I will then reinterpret this problem as bipartite matching, and discuss some recently-discovered implications of this interpretation.

It's worth noting that the transductive perspective is of perhaps limited utility for binary classification, where empirical risk minimization (ERM) has almost optimal PAC sample complexity~\cite{shalev-shwartz_understanding_2014}. It is in multiclass classification where transductive learners truly shine, in particular as the number of labels grows large. In fact, ERM can fail to learn entirely even for learnable multiclass problems~\cite{daniely_multiclass_2011}, as can SRM and any proper learner \cite{daniely_optimal_2014}.  To my knowledge, all known general-purpose learners for multiclass classification factor through the transductive model and  OIGs~\cite{daniely_optimal_2014, brukhim_characterization_2022,aden-ali_optimal_2023,asilis_regularization_2024}.

\subsection{One-Inclusion Graphs}
\label{sec:oig}

The \emph{one-inclusion graph (OIG)} is a mathematical representation of classification in the transductive model of learning. OIGs were proposed by  \citet{haussler_predicting_1994} to model  realizable binary classification, and have since been  extended to realizable multiclass classification by \citet{rubinstein_shifting_2009}, and  to the agnostic setting by~\citet{asilis_regularization_2024}. As we will see shortly, OIGs represent these learning tasks exactly, with their \emph{orientations} in one-to-one correspondence with transductive learners. 

\paragraph{Realizable Binary Classification.} I will start by describing OIGs in their original form, restricted to realizable binary classification. Fix a domain $\X$ and a hypothesis class $\H \sse \set{0,1}^\X$. Recall that, in the  realizable setting of the transductive model, a collection of $n$  unlabeled datapoints $S=(x_1,\ldots,x_n) \in \X^n$ is chosen by an adversary, and provided to the learner at the outset. The adversary also selects a \emph{ground truth} hypothesis $h^* \in \H$,  then the learner is asked to predict the label $h^*(x_i)$ of a randomly-chosen test datapoint $x_i$ from the labels $h^*|_{S_{-i}}$ of the remaining $n-1$ datapoints.

\begin{wrapfigure}{R}{0.42\textwidth}
  \centering
  \begin{tikzpicture}[vertex/.style={circle, draw, minimum size=20pt, inner sep=1pt}]

\node[vertex] (001) at (0,0) {(0,0,1)};
\node[vertex] (010) at (3,1) {(0,1,0)};
\node[vertex] (100) at (2,2) {(1,0,0)};
\node[vertex] (101) at (0,2) {(1,0,1)};
\node[vertex] (110) at (3,3) {(1,1,0)}; 
\node[vertex] (111) at (1,3) {(1,1,1)};

\draw (001) -- (101) -- (111) -- (110) -- (010) -- cycle;
\draw (100) -- (101);
\draw (110) -- (100);

\end{tikzpicture}

  \caption{Example OIG for realizable binary classification on three~datapoints.}
  \label{fig:binaryoig}
\end{wrapfigure}
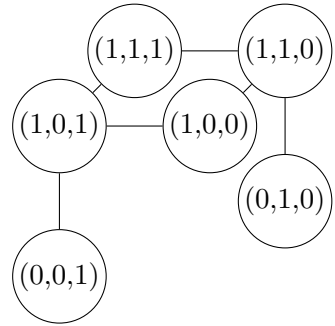
The OIG is an undirected graph $G(\H,S)$  determined by the hypothesis class $\H$ and the unlabeled datapoints  $S$, both of which are known to the learner. The nodes correspond to the possible labelings $\H|_S$  of the datapoints $S$ by hypotheses in $\H$, of which there can be at most $2^n$.\footnote{A tighter bound is given by the \emph{growth function} $\Phi_d(n) = \sum_{i=0}^d \binom{n}{i}$, where $d$ is the VC dimension of $\H$.} 
  An edge is included between two labelings if they differ on exactly a single datapoint. Another perspective is that the OIG is simply the subgraph of the $n$-dimensional hypercube induced by $\H|_S$. An example OIG is depicted in Figure~\ref{fig:binaryoig}.

  Now consider a learner trying to predict the label of $x_i$ from labels $\set{y_j}_{j \neq i}$ provided for all the other datapoints. If there is a single node $(y_1, \ldots, y_i, \ldots, y_n)$  in $G$ that is consistent with the provided labels, then there is nothing to decide: $y_i$ must be the correct label for $x_i$. The interesting case is when there are two such nodes, one of which labels $x_i$ with a $0$ and the other with a~$1$. These two nodes, in differing only on $x_i$, must share an edge $e$ in the OIG. In such a situation, a learner predicting a label for $x_i$ chooses one of the two nodes to ``go with'', effectively \emph{orienting} $e$   towards the chosen prediction. In specifying a prediction for every such scenario, a learner orients all the edges of the OIG, turning it into a directed graph.  
To summarize, learners are  in one-to-one correspondence with \emph{orientations} of the OIG.\footnote{More accurately, this is the case for deterministic learners. Randomized learners correspond to randomized orientations. This distinction is not especially consequential, so I blur it in this article.} 

Given this graph-theoretic view of a transductive learner, we can also describe its error guarantee  in graph-theoretic terms: it is proportional to the maximum \emph{out-degree} of a node in the oriented graph. To see this, observe that each node  represents a possible ground truth labeling $\y=(y_1, \ldots, y_n)$ of the datapoints, and each edge represents an ambiguity faced by the learner looking to predict some $y_i$. When an edge is directed out of $\y$, this corresponds to a ``mistake'' by the learner when $\y$ happens to be the ground truth. Since the test datapoint~$x_i$ is chosen uniformly at random, and we evaluate error in the worst case over possible ground truths, we can conclude that the learner's error (i.e., worst-case misclassification rate) equals the maximum outdegree divided by $n$. Since each edge results in a mistake for one of its endpoints, one can think of transductive learning as seeking to ``spread the error around'' as evenly as possible among all possible ground truths.

\paragraph{Realizable Multiclass Classification.} One-inclusion graphs generalize naturally to multiclass classification, as articulated by \cite{rubinstein_shifting_2009}. For a set $\Y$ of labels, which may be finite or infinite, the nodes of the OIG are still the  ground truth labelings $\y=(y_1,\ldots,y_n) \in \Y^n$ consistent with some hypothesis. The edges, however, turn into hyper-edges, with a collection of nodes sharing a hyper-edge precisely when they all disagree on \emph{the same} $y_i$, and no other.  An OIG involving  three datapoints and three labels  is depicted in Figure~\ref{fig:oig}. 
The hyper-edges again correspond to the ambiguities faced by the learner, where multiple predictions (up to $|\Y|$ many) are consistent with provided data. A learner can therefore be seen as orienting a hyper-edge towards the node corresponding to its prediction, and away from the others. As before, a ground truth incurs one ``mistake'' for every incident edge directed away from it.  Defining the out-degree of a node as the number of its incident edges directed away from it, the learner's error is again the maximum out-degree divided by $n$.

\definecolor{myblue}{RGB}{0,80,160}
\definecolor{mygreen}{RGB}{80,160,80}
\begin{figure}[tbp]
    \centering
    \begin{subfigure}[b]{0.45\textwidth}
        \centering
        \begin{tikzpicture}
    \tikzstyle{nodeStyle}=[circle, thick, draw=mygreen, minimum size=1.5em, inner sep=2pt]

    \tikzstyle{lineEdge}=[myblue, thick]

        \node[nodeStyle] (n0) at (0,2) {$(1, 0, 0)$};
    \node[nodeStyle] (n1) at (0,0) {$(0, 0, 0)$};
    \node[nodeStyle] (n2) at (2,0) {$(0, 1, 0)$};
    \node[nodeStyle] (n3) at (4,0) {$(0, 2, 0)$};
    \node[nodeStyle] (n4) at (4,2) {$(1, 2, 0)$};


    \node[draw=myblue, fit=(n1) (n2) (n3), inner sep=6pt, thick, rectangle, rounded corners] {};

    \node[draw=myblue, fit=(n0) (n1), inner sep=1pt, thick, rectangle, rounded corners] {};
    \node[draw=myblue, fit=(n3) (n4), inner sep=1pt, thick, rectangle, rounded corners] {};

\end{tikzpicture}

        \caption{Traditional OIG. Nodes depicted with circles,  hyper-edges with rectangles.}
        \label{fig:oig}
    \end{subfigure}
    \hfill 
    \begin{subfigure}[b]{0.45\textwidth}
        \centering
        \begin{tikzpicture}[thick,
  every node/.style={draw,circle},
  fsnode/.style={fill=myblue},
  ssnode/.style={fill=mygreen},
  every fit/.style={ellipse,draw,inner sep=-2pt,text width=1cm},
  ->,shorten >= 3pt,shorten <= 3pt, scale=0.5]

\begin{scope}[start chain=going below,node distance=4mm, yshift=-3em]
  \node[fsnode,on chain] (f0) [label=left: {$(?, 0, 0)$}]  {};
  \node[fsnode,on chain] (f1) [label=left: {$(0, ?, 0)$}]  {};
  \node[fsnode,on chain] (f2) [label=left: {$(?, 2, 0)$}] {};
\end{scope}

\begin{scope}[xshift=5 cm,start chain=going below,node distance=3mm]
  \node[ssnode,on chain] (s0) [label=right: {$(1, 0, 0)$}] {};
  \node[ssnode,on chain] (s1) [label=right: {$(0, 0, 0)$}] {};
  \node[ssnode,on chain] (s2) [label=right: {$(0, 1, 0)$}] {};
  \node[ssnode,on chain] (s3) [label=right: {$(0, 2, 0)$}] {};
  \node[ssnode,on chain] (s4) [label=right: {$(1, 2, 0)$}] {};
\end{scope}


\draw (f1) -- (s1);
\draw (f1) -- (s2);
\draw (f1) -- (s3);
\draw (f2) -- (s3);
\draw (f2) -- (s4);
\draw (f0) -- (s0);
\draw (f0) -- (s1);

\end{tikzpicture}

        \caption{Bipartite OIG. Left side nodes with a single neighbor, such as $(0,0,?)$, omitted for clarity.}
        \label{fig:bpoig}
    \end{subfigure}
    \caption{Example OIG for realizable multiclass classification, in traditional and bipartite forms.}
    \label{fig:bothoig}
\end{figure}
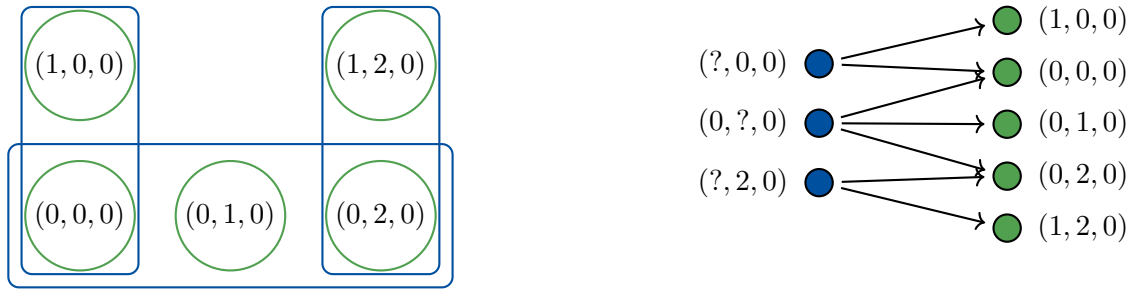


\paragraph{Agnostic Multiclass Classification.} 
OIGs were recently further generalized to the agnostic setting of multiclass classification by  \citet{asilis_regularization_2024}, building on a closely-related definition employed by \citet{long_complexity_1999} for a different model of learning. This \emph{agnostic OIG}, denoted $G^\ag(\H,S)$, is as follows.  Since the ground truth labeling of the given datapoints $S=(x_1,\ldots,x_n)$ may now be completely arbitrary, we include all labelings $\Y^n$ as nodes of $G^\ag$.  As in the realizable case, a collection of nodes share a hype-edge precisely when they disagree on the same $y_i$ and no other.   Also as in the realizable case, a learner is an orientation of the hyper-edges, and the learner's error is related to out-degrees of all the nodes.

But what of the hypothesis class $\H$? Since we are in the agnostic setting, $\H$ serves merely as a benchmark for our learner. Therefore, in defining the OIG orientation problem we now ``discount'' the out-degree of each node  $\y=(y_1,\ldots,y_n)$ of $G^\ag$ by its \emph{Hamming distance} $\dist(y,\H)$ from $\H|_S$.\footnote{The Hamming distance between two vectors is the number of entries on which they differ. For the distance to a set of vectors, we minimize over that set appropriately.} 
This is quite natural, as the more the ground truth deviates from our benchmark class $\H$, the more we are ``off the hook'' for making accurate predictions.

\paragraph{A Note on Computational Complexity.} Some remarks are in order with regard to the computational complexity of working with the OIG. Even for binary classification with a finite VC dimension $d$, the OIG can have on the order of $n^d$ nodes. This is prohibitively large for all but the smallest values of $d$. In multiclass classification the situation can be even more dire. In fact, there are learnable problems with infinitely many labels and  OIGs of infinite size! It is therefore unsurprising that OIGs have usually not been thought of as a practical data structure for learning, but rather as abstractions for its mathematical study. In some sense, orienting the OIG can be viewed as ``brute-forcing'' the learning task: a representation of the algorithm's entire input/output behavior, at least with respect to labelings, is computed before even looking at the labels in the training data!  This stands in contrast to practical approaches  such as ERM and SRM (see Section~\ref{sec:background}), 
which represent the task parsimoniously and reason locally, input by input. Nonetheless, I will argue in this article that insights emanating from OIGs, and extensions of them, can  point the way to such tenable algorithms.

\subsection{The Matching Perspective}
\label{subsec:matching}

One inclusion graphs transform multiclass learning to what is essentially a combinatorial optimization problem on orientations, albeit one that is described implicitly. From here, a simple shift in perspective articulated by \cite{asilis_regularization_2024}, employing the classical connection between orientation and matching problems (see e.g. \cite{Schrijver_2003}), yields an equivalent bipartite matching problem. I will describe this next.

Let $G$ be an OIG associated with hypothesis class $\H \sse \Y^\X$ and unlabeled dataset $S=(x_1,\ldots,x_n)$. The transformation is essentially identical whether we let $G$ be the realizable or agnostic OIG, but I recommend the reader keep the realizable OIG in mind to keep things simple. On the right we have the  nodes of $G$, corresponding to possible ground-truth labelings $\y=(y_1,\ldots,y_n)$ of  $(x_1,\ldots,x_n)$ . On the left we have \emph{partial labelings} which occlude a single label from some ground truth. We represent these as a vector $(y_1, \ldots, ?, \ldots, y_n)$,  where $y_i \in \Y$ for all but a single ``?'' entry corresponding to the occluded label. An edge is included between a partial labeling on the left and a full labeling on the right if they agree on all but occluded label. 
In graph-theoretic terms,  the bipartite OIG $G_{\bp}$ is essentially  the \emph{edge-vertex incidence graph} of $G$, were $G$ to first be augmented with self-loops (i.e., singleton hyper-edges) so that every node has degree exactly~$n$. 
A bipartite OIG for the realizable setting is depicted in Figure~\ref{fig:bpoig}.

The  left side nodes of $G_{\bp}$ --- the partial labelings --- represent the scenarios faced by the learner, where one label is omitted and the learner is asked to ``fill in the blank.'' Edges of $G_{\bp}$ encode the consistency relationship between the partial labelings on the left and the (full) labelings on the right. A learner, in having to predict each occluded label, can be viewed as an \emph{assignment} mapping each node on the left to one of its neighbors on the right. Note that each node on the right has degree exactly $n$ in $G_{\bp}$, one for each possible position of the ``?'',  whereas nodes on the left may have degree up to the number of labels $|\Y|$ (exactly $|\Y|$ in the agnostic setting). 

Given this perspective on learners as assignments in the bipartite OIG, what of their error? It is easier to take a complementary perspective and describe the learner's \emph{accuracy}: the probability it correctly classifies the test point, in the worst case over ground truths. For a particular ground truth labeling $\y=(y_1,\ldots,y_n)$ on the right side of $G_{\bp}$, the learner faces $n$ equally likely scenarios, one for each left node $\tilde{\y} = (y_1, \ldots, ? , \ldots, y_n)$ incident on $\y$. The learner correctly classifies the test point precisely when it assigns $\tilde{\y}$ to  $\y$. Therefore, viewing a learner as a left to right assignment in $G_{\bp}$, its accuracy is its minimum degree to a node on the right, divided by $n$.  In the realizable setting, the error  is of course the complementary probability. In the agnostic setting, we discount each degree of a node $\y$ on the right by its Hamming distance $\dist(\y,\H)$ to the hypothesis class, then proceed identically to arrive at the agnostic error. 

We can now cast learning  as a  bipartite matching problem. To obtain error $\epsilon$, each node $\y$ on the right side of $G_{\bp}$ must be assigned --- we say \emph{matched} --- at least $(1-\epsilon) n$ times in the realizable case, and at least $(1-\epsilon) n - \dist(\y,\H)$ times in the agnostic case. This is what is referred to in combinatorial optimization as a  bipartite \emph{$b$-matching problem}, a generalization of bipartite matching where nodes can be prescribed a minimum required number of matches. This generalization is exclusively for convenience, as a bipartite $b$-matching problem can be equivalently posed as  a run-of-the-mill  maximum bipartite matching problem by appropriately cloning nodes that require multiple matches. It must be noted, however, that $G_{\bp}$ can like $G$ be an  infinite graph, and even when finite is typically prohibitively large for invoking matching algorithms directly.

What, then, does this matching perspective on classification  buy us? Quite a bit, as it turns out. I will describe some recent consequences  of this viewpoint next.

\subsection{Implication of Matching: A Compact Graph-Theoretic Characterization}
\label{sec:hall}

Characterizing the optimal sample complexity of  learning  is a chief concern of learning theory. Given the simplicity of the transductive model, one might hope for a crisp and interpretable expression for the sample complexity, or equivalently the error rate, of the optimal transductive learner. Such an expression would approximate the performance of the optimal PAC learner, as per the relationships described  in Section \ref{subsec:pac_trans}. The matching perspective yields such a graph-theoretic expression which captures the transductive  error rate ``on the nose'' \cite{asilis_regularization_2024}. Moreover, this expression refers only to  \emph{finite projections} of the learning problem.

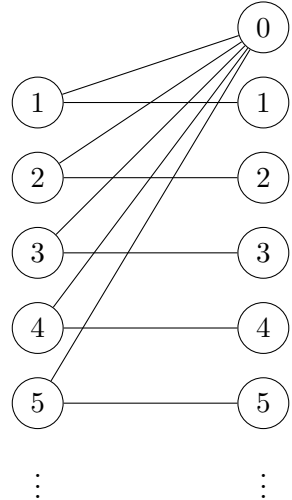
\begin{wrapfigure}{R}{0.41\textwidth}
  \centering
  \scalebox{1}{\begin{tikzpicture}

\foreach \i in {1,...,5} {
    \node[circle, draw, fill=white] (L\i) at (0, -\i) { \i };
    \node[circle, draw, fill=white] (R\i) at (3, -\i) { \i };
}

\node[circle, draw, fill=white] (R0) at (3, 0) { 0 };

\foreach \i in {1,...,5} {
    \draw[-] (L\i) -- (R\i);
    \draw[-] (L\i) -- (R0);  
}

\node at (0, -6) {\vdots};
\node at (3, -6) {\vdots};

\end{tikzpicture}
}
  \caption{Matching on infinite bipartite graphs absent degree constraints.}
  \label{fig:infinite_matching}
\end{wrapfigure}
The starting point here is Philip Hall's classic \emph{marriage theorem}~\cite{hall1987representatives}, which characterizes exactly when one side of a \emph{finite} bipartite graph can be matched to the other. Let $L$ and $R$ denote the left and right side nodes of the graph, respectively, and consider whether there is a matching which matches every node on the right. The original marriage theorem states that such a matching exists  if and only if every set $R' \sse R$ of right side nodes has at least $|R'|$ neighbors between them (on the left, naturally). Whereas this \emph{Hall condition} characterizes matching on finite bipartite graphs, it fails to extend to all infinite bipartite graphs, where it is necessary but not sufficient in general. This can be seen by way of a simple example, illustrated in Figure~\ref{fig:infinite_matching}.

The ``troublemaker'' in Figure~\ref{fig:infinite_matching} is the infinite-degree node $0$ on the right. If we assume that the nodes we intend to match --- the right side nodes $R$, in our discussion --- have finite degrees, the marriage theorem can be recovered:  we can match  $R$ if and only if we can match every finite $R' \sse R$, which in turn is possible precisely when each finite $R' \sse R$ has at least $|R'|$ neighbors (the Hall condition). This holds for infinite graphs of arbitrary, even uncountable, cardinality, so long as the degrees on the right are all finite (not necessarily bounded); degrees on the left may be arbitrary, even uncountable. This \emph{infinite marriage theorem} was first established by Marshall Hall~\cite{jr_distinct_1948} by way of a quite sophisticated proof,\footnote{No relation to Philip Hall, coincidentally.}  though somewhat simpler proofs have since been discovered (e.g. \cite{halmos_marriage_1950,rado_note_1967}). The infinite marriage theorem also extends naturally to  $b$-matching by a simple cloning argument, under the same requirement of finite right-side degrees: If each node on the left can be matched at most once, whereas each node $r$ on the right requires at least $b_r$ matches, then this is possible if and only if it is possible for every finite $R' \sse R$, which in turn is possible if and only if each finite $R' \sse R$ has at least $b(R') = \sum_{r \in R'} b_r$ neighbors. In showing that an infinite graph admits a matching of a particular form precisely when the same holds for some of its finite subgraphs, M. Hall's infinite marriage theorem is what is referred to as a \emph{compactness} result.

This now brings us naturally to bipartite OIGs, which encode classification as a matching problem. In the realizable setting, a learner with error $\epsilon$ corresponds to a $b$-matching on $G_{\bp}$ with $b=(1-\epsilon)n$ for every right side side node $\y=(y_1, \ldots, y_n)$. The infinite marriage theorem then implies that this is possible precisely when every finite family $R'$ of full labelings on the right side of $G_{\bp}$ is incident on at least $(1-\epsilon)n \cdot |R'|$ partial labelings. This yields a precise expression of the optimal error as a function of the sample size $n$,  as the worst-case  ``matchability'' of finite subgraphs of a bipartite OIG. This graph-theoretic expression, stated precisely in \cite{asilis_regularization_2024}, is called the \emph{Hall complexity} of the hypothesis class. The expression is easily adapted to the agnostic setting by appropriately discounting the matching requirement of each node $\y$ in the agnostic OIG by $\dist(\y,\H)$. 

Much like in the case of matching, we now have a compactness result for classification. The Hall complexity expresses the optimal transductive error rate of a classification problem as the worst-case learning rate of its \emph{finite projections}: those problems which can be obtained by restricting attention to finite subsets of the domain $\X$ and finite subsets $\H' \sse \H$ of the hypothesis space. To see this, note that  each bipartite OIG is a function of a finite subdomain $x_1,\ldots,x_n$, and a finite subset of its right side nodes corresponds to finitely many hypotheses.  By exactly relating the matchability in the bipartite OIG to the same matchability in some of its finite subgraphs, we can conclude that the maximal obstructions to transductive classification are finite in nature. 

\paragraph{Bibliographic Remarks.} Simple and \emph{compact} expressions for the optimal error, or equivalently the sample complexity, have long been sought in learning theory. For realizable binary classification, the sample complexity is characterized in terms of the VC dimension both in the transductive \cite{haussler_predicting_1994,li_one-inclusion_2001} and PAC (see e.g. \cite{shalev-shwartz_understanding_2014}) models.    For multiclass classification, the \emph{DS dimension} of \citet{daniely_optimal_2014} yields an approximation of the optimal error in the realizable setting up to a constant in the exponent, as was recently shown by \citet{brukhim_characterization_2022}. Outside of  \emph{dimensions}, a quantity derived from the (classical) one-inclusion graph $G$ gives a constant factor approximation: the maximum average degree of a subgraph of $G$, divided by $n$, is with a factor of~$2$ of the optimal error in the realizable setting. The Hall complexity can be viewed as ``sharpening'' the maximum-average degree into an exact characterization, and extending it to the agnostic setting.

The most sought-after characterizations of learning are \emph{dimensions} such as the VC dimension, DS dimension, and others. Intuitively, a dimension measures the size of the ``largest obstruction'' to learning, with an infinite dimension implying unlearnability, and a finite dimension yielding a bound on the sample complexity by way of a simple expression. Compactness results of the sort provided by the Hall complexity are a weaker requirement than the existence of a dimension capturing optimal learning. I will elaborate on the relationship between compactness and dimensions in Section~\ref{sec:compactness}.  

\subsection{Implication of Matching: An Algorithmic Template for Classification}
Another important concern of learning theory is identifying algorithmic approaches that are broadly successful for a large swath of learning problems, and moreover simple enough to describe, implement, and explore. Perhaps the best example of this is empirical risk minimization (ERM), as described in Section~\ref{sec:background}. ERM and closely related approaches such as structural risk minimization (SRM) pervade the practice of machine learning, and in some cases are also accompanied by theoretical guarantees.  Most notably, for binary classification problems in the PAC model, it has long been known that ERM characterizes learnability and moreover has almost-optimal sample complexity (see e.g. \cite{shalev-shwartz_understanding_2014}).

Such theoretical optimality guarantees  for ERM  tend to fall by the wayside as one moves beyond binary classification.  In fact, ERM can entirely fail to learn in some settings where learning is possible, as shown by \citet{shalev-shwartz_learnability_2010}.   This failure manifests   in multiclass classification as shown by \citet{daniely_multiclass_2011}, and in a slight generalization of binary classification as shown by \citet{alon_theory_2021}. Even in bounded agnostic linear regression, the sample complexity of ERM was recently shown to be quite suboptimal by \citet{vaskevicius_suboptimality_2023}.\footnote{That said, ERM characterizes learnability for bounded regression  in the coarse sense, disregarding sample complexity~\cite{alon_scale-sensitive_1997}.}  

In settings where ERM fails to characterize optimal learning, the algorithms which do succeed in theory are often far removed from any parsimonious description or any heuristic employed in practice.  This is most apparent in multiclass classification, where the known near-optimal algorithms \cite{daniely_optimal_2014,brukhim_characterization_2022} involve explicitly-described orientations of the one-inclusion graph. Can the bipartite matching interpretation simplify these OIG-based algorithms? Bipartite matching tends to respond well to ``simple'' heuristics such as greedy algorithms, primal-dual approaches, and others --- can the same hold true for learning?

 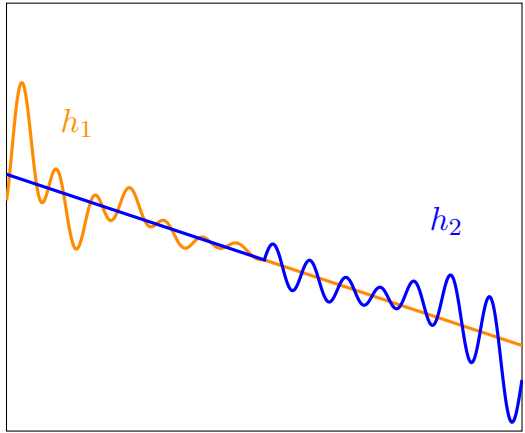
\begin{figure}
\centering
\begin{tikzpicture}[define rgb/.code={\definecolor{mycolor}{RGB}{#1}},
    rgb color/.style={define rgb={#1},mycolor}, scale=1.0]
\begin{axis}[
    xmin = -15, xmax = 15,
    ymin = -2.0, ymax = 3.0,
    ticks=none]
    \addplot[
        domain = -15:0,
        samples = 200,
        smooth,
        very thick,
        rgb color = {255, 140, 0}
    ] {-x/15 + exp(-x / 5) * sin(deg(x)) * 1/15 * ( cos(deg(x) * 2) +  0.2 * sin(deg(x)*2) )}
    node [pos=0.3, above right] [label={[label distance=0.5cm]above: {\scalebox{1.15}{$h_1$}}}]{}; 
    \addplot[
        domain = 0:15,
        samples = 200,
        smooth,
        thick,
        very thick,
        rgb color = {255, 140, 0}
    ] {-x/15}; 
    \addplot[
        domain = 0:15,
        samples = 200,
        smooth,
        very thick,
        rgb color = {0, 0, 255}
    ] {-x/15 + exp(x / 5) * sin(deg(x)) * 1/20 * ( cos(deg(x) * 1.5)) +  0.2 * sin(deg(x)*3) }
    node [pos=0.7, above left] [label={[label distance=0.4cm]above: {\scalebox{1.15}{$h_2$}}}]{}; 
    \addplot[
        domain = -15:0.1,
        samples = 200,
        smooth,
        very thick,
        rgb color = {0, 0, 255}
    ] {-x/15};
\end{axis}
\end{tikzpicture}

\caption{\emph{ 
    A  local regularizer may favor the simplicity of $h_1$ on test points drawn from the right region of the domain, and the simplicity of $h_2$ on test points drawn from the left region.}}
\label{fig:local}
\end{figure}

The answer to both questions turns out to be yes, as shown recently by~\citet{asilis_regularization_2024}. Through primal-dual analysis of a certain convex program over matchings in the bipartite OIG, we derive a near-optimal learner for multiclass classification of a particular instructive form. In particular, this learner implements a variation of the SRM template from Section~\ref{sec:background}, relaxed on two fronts: (1) It employs  a  \emph{local} regularization function  $\psi(h,x_{\tst})$ that  measures predictor complexity differently depending on the test point, and  (2) it incorporates an unsupervised pre-training stage which learns this regularization function.

These relaxed SRMs --- derived from the connection to bipartite matching --- provide a  simpler template for optimal learning which is quite natural, and is reminiscent of approaches discovered to be useful in practice. Relaxing to local regularization is well-motivated, since different hypotheses may be simple in some areas of the domain and complex in others; this is illustrated in Figure~\ref{fig:local}. It is therefore not altogether surprising that this result is predated by applications of local regularization to image classification and restoration~\cite{wolf_local_2008,prost_learning_2021}. The second relaxation is in line with much of the recent empirical and theoretical evidence on the utility of an unsupervised learning stage. Indeed, unsupervised pre-training has seen widespread application to  computer vision, natural language processing, and speech recognition (see the discussion in \cite{ge_provable_2024}).


\newcommand{\fds}{FDS\xspace}
\newcommand{\fdss}{FDSs\xspace}

\section{Learning with General Loss Functions}
\label{sec:general}
In this section I will show how the bipartite matching perspective, appropriately generalized, can lead to insights for supervised learning beyond classification. First I describe a generalization of the bipartite OIG, implicit in \citet{asilis_transductive_2024}, intended to capture learning beyond classification. I then overview one application of this approach, also from \cite{asilis_transductive_2024}: Appropriately extending Hall's theorem leads to a quite-general \emph{compactness result} for learning, one which relates the sample complexity of a problem to that of its finite projections, in both the transductive and PAC models.

\subsection{A Generalization of the One-Inclusion Graph}
\label{sec:fds}


A \emph{Functional Dependency Structure (\fds)} is a potentially-infinite bipartite graph with \emph{input variables}   on the left, \emph{output variables}  on the right, and edges annotated with functions encoding the dependencies between inputs and outputs. Let $A$ and $B$ be index sets for the left and right side nodes of the graph, respectively, and let $E \sse A \cross B$ be the set of its edges. For each left side node $a \in A$ we associate an input variable $u_a$ which takes values in some set $\Z_a$. For each right side node $b \in B$ we associate an output variable $v_b$ which takes real number values. Each output variable depends on finitely many input variables, and we include the edge $(a,b)$ in $E$ whenever $v_b$ depends on $u_a$.  Notably, nodes on the right have finite degrees, but we make no such assumption for the degrees of nodes on the left.
Each edge $e=(a,b)$ is labeled with a \emph{cost function} $f_{e}: \Z_a \to \RR$, and the output variables average the incident cost functions. More concretely,  $v_b$ evaluates to the average, over all its incident edges $e=(a,b)$,  of $f_{e}(u_a)$. 

Given an \fds, our objective is to assign values to the input variables so as to minimize the maximum output variable. If an assignment to the inputs maintains all outputs  weakly below $\epsilon$, we call it an \emph{$\epsilon$-assignment}.
%
%
Next, I describe how this \emph{\fds assignment problem} generalizes both bipartite matching and transductive learning with general losses.

\paragraph{Infinite Bipartite Matching.}  Consider  matching one side of an infinite bipartite graph into the other, where the side  to be matched (say, the right side) has finite degrees. Let $G$ be a bipartite graph with left side $A$, right side $B$, and edges $E$. Let $\deg(b)$ denote the degree of a node $b \in B$, which we assume is finite.  We associate with each $a \in A$ a variable $u_a$ with domain $\Z_a \sse B$ equal to the neighbor set of $a$. For each edge $e=(a,b) \in E$, we associate the cost function $f_e(z) = [z \neq b] + \frac{1}{\deg(b)}$. A simple calculation shows that a matching of $B$ into $A$ corresponds to a $1$-assignment in this \fds. 

\paragraph{Classification.} As a warm-up, I will start with classification in the realizable setting. Given a hypothesis class  $\H \sse \Y^\X$ and  datapoints $S=(x_1,\ldots,x_n) \in \X^n$,  the bipartite OIG $G_{\bp}(\H,S)$ can easily be seen as an \fds.  For each partial labeling $\tilde{\y}=(y_1,\ldots, ?, \ldots, y_n)$ on the left side of $G_{\bp}$,  with ``?'' in the $i$th entry, we associate an input variable $u_{\tilde{\y}}$ with domain $\Y$. An assignment to $u_{\tilde{y}}$ is interpreted as a prediction for  $y_i$ when the learner is given the remaining labels $\set{y_j}_{j \neq i}$. For each full labeling $\y=(y_1,\ldots,y_n)$ on the right side of $G_{\bp}$ we associate an output variable $v_{\y}$, which measures the misclassification rate in  ground truth $\y$. The edges $(\tilde{\y},\y)$ of $G_{\bp}$, if labeled with $f_{\tilde{y},y}(z) = [z \neq y_i]$ when $i$ is the index of ``?'' in $\tilde{y}$, now appropriately encode the dependencies between input and output variables. Notice that the nodes on the right have finite degree $n$, as required in an \fds. An $\epsilon$-assignment for this \fds now corresponds to a  transductive learner with realizable error at most $\epsilon$.

For agnostic classification, the bipartite agnostic OIG $G_{\bp}^{\ag}(\H,S)$ can be similarly interpreted as an \fds, with one main difference: Since we benchmark our learner relative to $\H$, we appropriately ``discount'' each cost function  by the best-in-class loss. Specifically, we let $f_{\tilde{y},y}(z) = [z \neq y_i] - \min_{h \in \H} \frac{1}{n} \sum_{j=1}^n  [h(x_j)~\neq~y_j]$. An $\epsilon$-assignment for this \fds now corresponds to a transductive learner with agnostic error at most $\epsilon$.

\paragraph{Learning with General Losses.} 
More generally, consider a supervised learning problem in the transductive model with hypotheses $\H \sse \Y^\X$, loss function $\ell: \Y \times \Y \to \RR_{\geq 0}$, and datapoints $S=(x_1,\ldots,x_n) \in \X^n$. Whether in the realizable or agnostic setting, we use the same \fds as in the corresponding classification problem except for appropriately exchanging the loss functions. Specifically, for $\y=(y_1,\ldots,y_n)$ and $\tilde{\y}=(y_1,\ldots, ?, \ldots y_n)$ with ``?'' at the $i$th position, we let $f_{\tilde{y},y}(z) = \ell(z,y_i)$ for the realizable setting, and $f_{\tilde{y},y}(z) = \ell(z,y_i) - \inf_{h \in \H} \frac{1}{n} \sum_{j=1}^n \ell(h(x_j),y_j)$ in the agnostic setting.  In either setting, an $\epsilon$-assignment for the appropriate \fds now corresponds to a transductive learner with error at most $\epsilon$.

\subsection{Implication: Compactness of Learning}
\label{sec:compactness}

 \emph{Compactness} refers to situations where a property of a mathematical object is determined by finite parts of that object. For example, an infinite graph is $k$-colorable if and only if the same holds for its finite subgraphs~\cite{bruijn1951colour}, a set of formulae in propositional logic is satisfiable precisely when any finite sub-family is satisfiable, and so on. Given that learning problems often feature infinite data domains, label sets, and hypothesis classes, it is natural to ask whether the same holds for learning. Specifically, can the difficulty of a learning problem, as measured by its sample complexity, be determined by inspecting its \emph{finite projections}? By those, recall that we mean the learning problems obtained by restricting to a finite set of data points, labels, and hypotheses.

Learning in the transductive model can be encoded exactly as an \fds assignment problem, as described in Section~\ref{sec:fds}.   The \fds assignment problem generalizes infinite bipartite matching problems of the form described in Section~\ref{sec:fds}, and those exhibit compactness via M. Hall's infinite marriage theorem~\cite{jr_distinct_1948} as discussed in Section~\ref{sec:hall}. Generalizing the infinite marriage theorem from bipartite graphs to \fdss would, therefore, extend compactness to learning.

This is precisely what we show in \citet{asilis_transductive_2024} by combining and extending ideas from three distinct proofs of the infinite marriage theorem \cite{jr_distinct_1948,rado_note_1967,halmos_marriage_1950}. Subject to mild topological assumptions on the input domains $\Z_a$ and cost functions $f_e$,\footnote{It suffices to assume that each $\Z_a$ has the topological structure of a metric space, and that each $f_e$ reflects compact sets, i.e.,  $f_e^{-1}(C)$ is compact whenever $C$ is compact.} an \fds admits an $\epsilon$-assignment precisely when the same holds for its subgraphs that are induced by a finite collection of right side nodes along with their neighbors. In other words, having an $\epsilon$-assignment is a compact property, being determined by the finite subgraphs of the \fds.  

Compactness of the \fds assignment problem implies a similar compactness result for most natural learning problems, including those with metric losses satisfying a mild topological assumption,\footnote{The metric space should satisfy the \emph{Heine-Borel} property, meaning that closed and bounded sets are compact.} as well as those with continuous losses on a compact label space. Such a problem is learnable with sample complexity $m(\epsilon)$ in the transductive model precisely when the same holds for all its finite projections. This holds for both the realizable and agnostic settings. 
These compactness results extend approximately to the PAC model, as per the relationships in Section~\ref{subsec:pac_trans}.

One interpretation of these results is the following: When learners are allowed to be improper, the obstacles to learning are finite in nature. This stands in contrast to influential recent work of \citet{ben-david_learnability_2019} which showed a different state of affairs when learners are constrained to be proper. There, a problem can be unlearnable --- or, even worse, its learnability may be independent from the axioms of ZFC set theory --- even though its finite projections are easily learned.\footnote{Note that \cite{ben-david_learnability_2019} do not pose their non-compact problem, which they term \emph{EMX learning}, as a supervised learning problem. However, it is not difficult to see that it can be rephrased as such.}

\paragraph{Compactness vs Dimensions.} Compactness seems to be a weaker property than the existence of a \emph{dimension} for learning, in the usual sense of the term. Dimensions such as the VC dimension, DS dimension, Natarajan dimension, and others are typically defined to equal the largest (finite) number of datapoints over which finitely many hypotheses are ``sufficiently rich.'' For problems with continuous loss functions, the notion of sufficiently rich can be scale sensitive, being parametrized by the desired error; a typical example of this is the fat shattering dimension for regression.  Whether scale sensitive or not, such a dimension is a  \emph{compact} function of the hypothesis class, and therefore so is any bound on the sample complexity purely in terms of that dimension. 

\section{Future Directions and Closing Thoughts}
I will have accomplished my main goal if at least some readers come away from this article thinking that the bipartite matching perspective on learning is at least sometimes useful. If not, then I will settle for the secondary goal of fostering a deeper appreciation --- among the uninitiated, at least --- of the transductive model and one-inclusion graphs as a lens into learning more broadly. Now I will take some time to speculate, perhaps irresponsibly, about possible future work that may benefit from the perspectives discussed in this article. 


\paragraph{Transductive vs PAC learning.} As described in Section~\ref{sec:trans}, transductive and PAC sample complexities are essentially equivalent (up to low-order terms) in the realizable setting, but may yet be separated by a multiplicative gap of up to $\frac{1}{\epsilon}$ in the agnostic setting. Can this gap be closed? In other words, are the transductive and PAC models also essentially equivalent in the agnostic setting of learning? 

\paragraph{Hat Puzzles and Learning.} The connection between hat puzzles and the transductive model, while easily recognizable as described in Section~\ref{sec:trans}, appears to not have been explored in the literature. Can ideas and techniques from one lead to progress in the other? 

\paragraph{Local Computation and Transductive Learning.} Classification in the transductive model can be viewed as a bipartite matching problem, as described in Section~\ref{sec:class}. This does not imply computational tractability, however, since the associated bipartite graph is exponentially large. Efficient learning here entails computing a node's match by examining only its local neighborhood, without constructing the entire graph. An algorithm precisely of this sort, for binary classification problems with an efficient ERM oracle, was recently shown by \citet{daskalakis_is_2024}. More generally, this is precisely the sort of problem tackled by the literature on \emph{local computation}, which was initiated by the work of \citet{rubinfeld_fast_2011}. Much work in that area has gone into designing local approximation algorithms for matching problems (e.g. \cite{kapralov_space_2020,levi_local_2015}). Can such local matching algorithms lead to computationally-efficient and near-optimal transductive learners? While this approach can not succeed in general due to hardness results for improper learning (e.g. \cite{daniely_local_2021,tiegel_improved_2024}), it may yet be plausible for large classes of  problems or under oracle assumptions as in~\cite{daskalakis_is_2024}.  

\paragraph{Generalizing Hall Complexity.} The Hall complexity (Section~\ref{sec:class}) characterizes the sample complexity of multiclass classification ``on the nose'' in the transductive model, and close to it in the PAC model. For more general loss functions, the closest such quantity I am aware of in the realizable setting is the $\gamma$-OIG dimension of \citet{attias_optimal_2023}, though there is substantial slack on the associated bounds on sample complexity. It seems quite plausible that a natural measure of complexity which does not qualify as a dimension, much like Hall complexity, can more tightly characterize sample complexity in large classes of (realizable and agnostic) learning problems with general loss functions.\footnote{One might be tempted to try the Rademacher complexity, covering number, or other quantities closely related to uniform convergence. However, those fail to characterize learnability even for multiclass classification \cite{daniely_multiclass_2011}.} Since the Hall complexity can be viewed as dual to the optimization problem in multiclass classification, perhaps a dual perspective on the \fds assignment problem from Section~\ref{sec:general} would yield such a measure?

\paragraph{Templates for Learning.} Local regularization and unsupervised pre-training, taken together, characterize optimal multi-class learning as described in Section~\ref{sec:class}.  Two questions stand out here: (a) Does this same template characterize near-optimal learning for problems beyond classification? Generalizing the analysis from bipartite OIGs to \fdss seems plausible as a route to such a result.  (b) Whether the unsupervised component is necessary remains an open question as described in~\citet{asilis_open_2024}: Can local regularization alone serve as a blueprint for optimal learning in classification, or even more generally?

\paragraph{Compactness vs Dimensions.} What is the exact relationship between compactness and dimensions? 
Could a learning problem exhibit compactness despite the lack of a dimension which characterizes learning, whether qualitatively or quantitatively? While I am not aware of any definitive answers, such a separation of compactness and dimensional characterizations  certainly seems plausible. As evidence for this on the quantitative front, the DS dimension for multiclass classification and the $\gamma$-OIG dimension for realizable regression --- both representing the state-of-the-art dimensions in their respective domains --- feature considerable slack in their characterization of sample complexity, despite exact compactness holding. Qualitative and quantitative separation of the two notions seems especially plausible for \emph{distribution family learning} problems, which make more nuanced distributional assumptions than the realizable or agnostic settings. The recent work of \citet{lechner_inherent_2024}, which  rules out dimensional characterizations for  some problems in distribution family learning,   might shed light on this~question.\footnote{Whereas \cite{asilis_transductive_2024} extends compactness to some particularly ``well-behaved''   distribution family learning problems, those appear to be disjoint from the problems covered by the impossibility results of \cite{lechner_inherent_2024}.}



\section*{Acknowledgments}
Many of the ideas in this article came out of conversations and joint papers  with a stellar group of collaborators. In alphabetical order, these are: Julian Asilis, Siddartha Devic, Yusuf Kalayci, Vatsal Sharan, Shang-Hua Teng, and Grayson York. The analogy between the transductive model and hat puzzles came out of conversations with Siyu Zeng.

{
\bibliography{learning,matching,misc,local_computation}

\begin{thebibliography}{50}
\providecommand{\natexlab}[1]{#1}
\providecommand{\url}[1]{\texttt{#1}}
\expandafter\ifx\csname urlstyle\endcsname\relax
  \providecommand{\doi}[1]{doi: #1}\else
  \providecommand{\doi}{doi: \begingroup \urlstyle{rm}\Url}\fi

\bibitem[Aden-Ali et~al.(2023)Aden-Ali, Cherapanamjeri, Shetty, and
  Zhivotovskiy]{aden-ali_optimal_2023}
I.~Aden-Ali, Y.~Cherapanamjeri, A.~Shetty, and N.~Zhivotovskiy.
\newblock Optimal {PAC} {Bounds} without {Uniform} {Convergence}.
\newblock In \emph{Proceedings of the 64th {Annual} {Symposium} on
  {Foundations} of {Computer} {Science} ({FOCS})}, pages 1203--1223. IEEE
  Computer Society, Nov. 2023.
\newblock ISBN 9798350318944.
\newblock \doi{10.1109/FOCS57990.2023.00071}.
\newblock URL
  \url{https://www.computer.org/csdl/proceedings-article/focs/2023/189400b203/1T971Rnelzi}.

\bibitem[Alon et~al.(1987)Alon, Haussler, and Welzl]{alon_partitioning_1987}
N.~Alon, D.~Haussler, and E.~Welzl.
\newblock Partitioning and geometric embedding of range spaces of finite
  {Vapnik}-{Chervonenkis} dimension.
\newblock In \emph{Proceedings of the third annual symposium on {Computational}
  geometry}, {SCG} '87, pages 331--340, New York, NY, USA, Oct. 1987.
  Association for Computing Machinery.
\newblock ISBN 978-0-89791-231-0.
\newblock \doi{10.1145/41958.41994}.
\newblock URL \url{https://dl.acm.org/doi/10.1145/41958.41994}.

\bibitem[Alon et~al.(1997)Alon, Ben-David, Cesa-Bianchi, and
  Haussler]{alon_scale-sensitive_1997}
N.~Alon, S.~Ben-David, N.~Cesa-Bianchi, and D.~Haussler.
\newblock Scale-sensitive dimensions, uniform convergence, and learnability.
\newblock \emph{Journal of the ACM}, 44\penalty0 (4):\penalty0 615--631, July
  1997.
\newblock ISSN 0004-5411.
\newblock \doi{10.1145/263867.263927}.
\newblock URL \url{https://dl.acm.org/doi/10.1145/263867.263927}.

\bibitem[Alon et~al.(2021)Alon, Hanneke, Holzman, and Moran]{alon_theory_2021}
N.~Alon, S.~Hanneke, R.~Holzman, and S.~Moran.
\newblock A {Theory} of {PAC} {Learnability} of {Partial} {Concept} {Classes}.
\newblock In \emph{Proceedings of the 62nd {IEEE} {Annual} {Symposium} on
  {Foundations} of {Computer} {Science} ({FOCS})}, pages 658--671. IEEE, 2021.

\bibitem[Anthony and Bartlett(1999)]{anthony_neural_1999}
M.~Anthony and P.~L. Bartlett.
\newblock \emph{Neural {Network} {Learning}: {Theoretical} {Foundations}}.
\newblock Cambridge University Press, Cambridge, 1999.
\newblock ISBN 978-0-521-57353-5.
\newblock \doi{10.1017/CBO9780511624216}.
\newblock URL
  \url{https://www.cambridge.org/core/books/neural-network-learning/665C8C7EB5E2ABC5367A55ADB04E2866}.

\bibitem[Asilis et~al.(2024{\natexlab{a}})Asilis, Devic, Dughmi, Sharan, and
  Teng]{asilis_open_2024}
J.~Asilis, S.~Devic, S.~Dughmi, V.~Sharan, and S.-H. Teng.
\newblock Open {Problem}: {Can} {Local} {Regularization} {Learn} {All}
  {Multiclass} {Problems}?
\newblock In \emph{Proceedings of {Thirty} {Seventh} {Conference} on {Learning}
  {Theory}}, pages 5301--5305. PMLR, June 2024{\natexlab{a}}.
\newblock URL \url{https://proceedings.mlr.press/v247/asilis24b.html}.
\newblock ISSN: 2640-3498.

\bibitem[Asilis et~al.(2024{\natexlab{b}})Asilis, Devic, Dughmi, Sharan, and
  Teng]{asilis_regularization_2024}
J.~Asilis, S.~Devic, S.~Dughmi, V.~Sharan, and S.-H. Teng.
\newblock Regularization and {Optimal} {Multiclass} {Learning}.
\newblock In \emph{Proceedings of {Thirty} {Seventh} {Conference} on {Learning}
  {Theory}}, pages 260--310. PMLR, June 2024{\natexlab{b}}.
\newblock URL \url{https://proceedings.mlr.press/v247/asilis24a.html}.
\newblock ISSN: 2640-3498.

\bibitem[Asilis et~al.(2024{\natexlab{c}})Asilis, Devic, Dughmi, Sharan, and
  Teng]{asilis_transductive_2024}
J.~Asilis, S.~Devic, S.~Dughmi, V.~Sharan, and S.-H. Teng.
\newblock Transductive {Learning} is {Compact}.
\newblock In \emph{The {Thirty}-eighth {Annual} {Conference} on {Neural}
  {Information} {Processing} {Systems}}, 2024{\natexlab{c}}.
\newblock URL \url{https://openreview.net/forum?id=YWTpmLktMj}.

\bibitem[Attias et~al.(2023)Attias, Hanneke, Kalavasis, Karbasi, and
  Velegkas]{attias_optimal_2023}
I.~Attias, S.~Hanneke, A.~Kalavasis, A.~Karbasi, and G.~Velegkas.
\newblock Optimal {Learners} for {Realizable} {Regression}: {PAC} {Learning}
  and {Online} {Learning}, Oct. 2023.
\newblock URL \url{http://arxiv.org/abs/2307.03848}.
\newblock arXiv:2307.03848 [cs, stat].

\bibitem[Ben-David et~al.(2019)Ben-David, Hrubeš, Moran, Shpilka, and
  Yehudayoff]{ben-david_learnability_2019}
S.~Ben-David, P.~Hrubeš, S.~Moran, A.~Shpilka, and A.~Yehudayoff.
\newblock Learnability can be undecidable.
\newblock \emph{Nature Machine Intelligence}, 1\penalty0 (1):\penalty0 44--48,
  Jan. 2019.
\newblock ISSN 2522-5839.
\newblock \doi{10.1038/s42256-018-0002-3}.
\newblock URL \url{https://www.nature.com/articles/s42256-018-0002-3}.
\newblock Number: 1 Publisher: Nature Publishing Group.

\bibitem[Bondy(1972)]{bondy_induced_1972}
J.~A. Bondy.
\newblock Induced subsets.
\newblock \emph{Journal of Combinatorial Theory, Series B}, 12\penalty0
  (2):\penalty0 201--202, Apr. 1972.
\newblock ISSN 0095-8956.
\newblock \doi{10.1016/0095-8956(72)90025-1}.
\newblock URL
  \url{https://www.sciencedirect.com/science/article/pii/0095895672900251}.

\bibitem[Bruijn and Erdos(1951)]{bruijn1951colour}
N.~d. Bruijn and P.~Erdos.
\newblock A colour problem for infinite graphs and a problem in the theory of
  relations.
\newblock \emph{Indigationes Mathematicae}, 13:\penalty0 371--373, 1951.

\bibitem[Brukhim et~al.(2022)Brukhim, Carmon, Dinur, Moran, and
  Yehudayoff]{brukhim_characterization_2022}
N.~Brukhim, D.~Carmon, I.~Dinur, S.~Moran, and A.~Yehudayoff.
\newblock A {Characterization} of {Multiclass} {Learnability}.
\newblock In \emph{Proceedings of the 63rd {Annual} {Symposium} on
  {Foundations} of {Computer} {Science} ({FOCS})}, pages 943--955, Denver, CO,
  USA, 2022. IEEE.
\newblock ISBN 978-1-66545-519-0.
\newblock \doi{10.1109/FOCS54457.2022.00093}.
\newblock URL \url{https://ieeexplore.ieee.org/document/9996893/}.

\bibitem[Butler et~al.(2009)Butler, Hajiaghayi, Kleinberg, and
  Leighton]{butler_hat_2009}
S.~Butler, M.~T. Hajiaghayi, R.~D. Kleinberg, and T.~Leighton.
\newblock Hat {Guessing} {Games}.
\newblock \emph{SIAM Review}, 51\penalty0 (2):\penalty0 399--413, May 2009.
\newblock ISSN 0036-1445, 1095-7200.
\newblock \doi{10.1137/080743470}.
\newblock URL \url{http://epubs.siam.org/doi/10.1137/080743470}.

\bibitem[Daniely and Shalev-Shwartz(2014)]{daniely_optimal_2014}
A.~Daniely and S.~Shalev-Shwartz.
\newblock Optimal learners for multiclass problems.
\newblock In \emph{Proceedings of the 27th {Conference} on {Learning} {Theory}
  ({COLT})}, pages 287--316. PMLR, 2014.

\bibitem[Daniely and Vardi(2021)]{daniely_local_2021}
A.~Daniely and G.~Vardi.
\newblock From {Local} {Pseudorandom} {Generators} to {Hardness} of {Learning}.
\newblock In \emph{Proceedings of 34th {Conference} on {Learning} {Theory}
  ({COLT})}, pages 1358--1394. PMLR, July 2021.
\newblock URL \url{https://proceedings.mlr.press/v134/daniely21a.html}.
\newblock ISSN: 2640-3498.

\bibitem[Daniely et~al.(2011)Daniely, Sabato, Ben-David, and
  Shalev-Shwartz]{daniely_multiclass_2011}
A.~Daniely, S.~Sabato, S.~Ben-David, and S.~Shalev-Shwartz.
\newblock Multiclass {Learnability} and the {ERM} principle.
\newblock In \emph{Proceedings of the 24th {Conference} on {Learning} {Theory}
  ({COLT})}, pages 207--232. JMLR Workshop and Conference Proceedings, 2011.
\newblock URL \url{https://proceedings.mlr.press/v19/daniely11a.html}.
\newblock ISSN: 1938-7228.

\bibitem[Daskalakis and Golowich(2024)]{daskalakis_is_2024}
C.~Daskalakis and N.~Golowich.
\newblock Is {Efficient} {PAC} {Learning} {Possible} with an {Oracle} {That}
  {Responds} '{Yes}' or '{No}'?, June 2024.
\newblock URL \url{http://arxiv.org/abs/2406.11667}.
\newblock arXiv:2406.11667 [cs, stat].

\bibitem[Dughmi et~al.(2024)Dughmi, Kalayci, and York]{dughmi_is_2024}
S.~Dughmi, Y.~Kalayci, and G.~York.
\newblock Is {Transductive} {Learning} {Equivalent} to {PAC} {Learning}?, May
  2024.
\newblock URL \url{http://arxiv.org/abs/2405.05190}.
\newblock arXiv:2405.05190 [cs, math, stat].

\bibitem[Ge et~al.(2024)Ge, Tang, Fan, and Jin]{ge_provable_2024}
J.~Ge, S.~Tang, J.~Fan, and C.~Jin.
\newblock On the {Provable} {Advantage} of {Unsupervised} {Pretraining}.
\newblock In \emph{Proceedings of the {Twelfth} {International} {Conference} on
  {Learning} {Representations} ({ICLR})}, 2024.
\newblock URL \url{https://openreview.net/forum?id=rmXXKxQpOR}.

\bibitem[Hall(1987)]{hall1987representatives}
P.~Hall.
\newblock On representatives of subsets.
\newblock \emph{Classic Papers in Combinatorics}, pages 58--62, 1987.

\bibitem[Hall~Jr(1948)]{jr_distinct_1948}
M.~Hall~Jr.
\newblock Distinct representatives of subsets.
\newblock \emph{Bulletin of the American Mathematical Society}, 54\penalty0
  (10):\penalty0 922--926, Oct. 1948.
\newblock ISSN 0002-9904, 1936-881X.
\newblock URL
  \url{https://projecteuclid.org/journals/bulletin-of-the-american-mathematical-society/volume-54/issue-10/Distinct-representatives-of-subsets/bams/1183512379.full}.
\newblock Publisher: American Mathematical Society.

\bibitem[Halmos and Vaughan(1950)]{halmos_marriage_1950}
P.~R. Halmos and H.~E. Vaughan.
\newblock The {Marriage} {Problem}.
\newblock \emph{American Journal of Mathematics}, 72\penalty0 (1):\penalty0
  214, Jan. 1950.
\newblock ISSN 00029327.
\newblock \doi{10.2307/2372148}.
\newblock URL \url{https://www.jstor.org/stable/2372148?origin=crossref}.

\bibitem[Haussler et~al.(1994)Haussler, Littlestone, and
  Warmuth]{haussler_predicting_1994}
D.~Haussler, N.~Littlestone, and M.~K. Warmuth.
\newblock Predicting \{0, 1\}-{Functions} on {Randomly} {Drawn} {Points}.
\newblock \emph{Information and Computation}, 115\penalty0 (2):\penalty0
  248--292, Dec. 1994.
\newblock ISSN 0890-5401.
\newblock \doi{10.1006/inco.1994.1097}.
\newblock URL
  \url{https://www.sciencedirect.com/science/article/pii/S0890540184710972}.

\bibitem[Kapralov et~al.(2020)Kapralov, Mitrović, Norouzi-Fard, and
  Tardos]{kapralov_space_2020}
M.~Kapralov, S.~Mitrović, A.~Norouzi-Fard, and J.~Tardos.
\newblock Space efficient approximation to maximum matching size from uniform
  edge samples.
\newblock In \emph{Proceedings of the {Thirty}-{First} {Annual} {ACM}-{SIAM}
  {Symposium} on {Discrete} {Algorithms} ({SODA})}, pages 1753--1772, USA,
  2020. Society for Industrial and Applied Mathematics.

\bibitem[Kearns and Vazirani(1994)]{kearns_introduction_1994}
M.~J. Kearns and U.~Vazirani.
\newblock \emph{An {Introduction} to {Computational} {Learning} {Theory}}.
\newblock MIT Press, Aug. 1994.
\newblock ISBN 978-0-262-11193-5.

\bibitem[Krzywkowski(2010)]{krzywkowski_hat_2010}
M.~Krzywkowski.
\newblock Hat problem on a graph.
\newblock \emph{Mathematica Pannonica}, 21:\penalty0 3--21, 2010.
\newblock URL
  \url{https://www.researchgate.net/profile/Marcin-Krzywkowski/publication/242731015_Hat_problem_on_a_graph/links/54ecf5900cf28f3e65351045/Hat-problem-on-a-graph.pdf}.

\bibitem[Lechner and Ben-David(2024)]{lechner_inherent_2024}
T.~Lechner and S.~Ben-David.
\newblock Inherent limitations of dimensions for characterizing learnability of
  distribution classes.
\newblock In \emph{Proceedings of {Thirty} {Seventh} {Conference} on {Learning}
  {Theory}}, pages 3353--3374. PMLR, June 2024.
\newblock URL \url{https://proceedings.mlr.press/v247/lechner24a.html}.
\newblock ISSN: 2640-3498.

\bibitem[Levi et~al.(2015)Levi, Rubinfeld, and Yodpinyanee]{levi_local_2015}
R.~Levi, R.~Rubinfeld, and A.~Yodpinyanee.
\newblock Local {Computation} {Algorithms} for {Graphs} of {Non}-{Constant}
  {Degrees}.
\newblock In \emph{Proceedings of the 27th {ACM} {Symposium} on {Parallelism}
  in {Algorithms} and {Architectures} ({SPAA})}, pages 59--61, New York, NY,
  USA, 2015. Association for Computing Machinery.
\newblock ISBN 978-1-4503-3588-1.
\newblock \doi{10.1145/2755573.2755615}.
\newblock URL \url{https://dl.acm.org/doi/10.1145/2755573.2755615}.

\bibitem[Li et~al.(2001)Li, Long, and Srinivasan]{li_one-inclusion_2001}
Y.~Li, P.~Long, and A.~Srinivasan.
\newblock The one-inclusion graph algorithm is near-optimal for the prediction
  model of learning.
\newblock \emph{IEEE Transactions on Information Theory}, 47\penalty0
  (3):\penalty0 1257--1261, Mar. 2001.
\newblock ISSN 1557-9654.
\newblock \doi{10.1109/18.915700}.
\newblock URL \url{https://ieeexplore.ieee.org/abstract/document/915700}.
\newblock Conference Name: IEEE Transactions on Information Theory.

\bibitem[Long(1999)]{long_complexity_1999}
P.~M. Long.
\newblock The {Complexity} of {Learning} {According} to {Two} {Models} of a
  {Drifting} {Environment}.
\newblock \emph{Machine Learning}, 37\penalty0 (3):\penalty0 337--354, Dec.
  1999.
\newblock ISSN 1573-0565.
\newblock \doi{10.1023/A:1007666507971}.
\newblock URL \url{https://doi.org/10.1023/A:1007666507971}.

\bibitem[Mohri et~al.(2018)Mohri, Rostamizadeh, and
  Talwalkar]{mohri_foundations_2018}
M.~Mohri, A.~Rostamizadeh, and A.~Talwalkar.
\newblock \emph{Foundations of {Machine} {Learning}, second edition}.
\newblock MIT Press, Dec. 2018.
\newblock ISBN 978-0-262-35136-2.

\bibitem[Montasser et~al.(2022{\natexlab{a}})Montasser, Hanneke, and
  Srebro]{montasser_adversarially_2022}
O.~Montasser, S.~Hanneke, and N.~Srebro.
\newblock Adversarially {Robust} {Learning}: {A} {Generic} {Minimax} {Optimal}
  {Learner} and {Characterization}, Sept. 2022{\natexlab{a}}.
\newblock URL \url{http://arxiv.org/abs/2209.07369}.
\newblock arXiv:2209.07369 [cs, stat].

\bibitem[Montasser et~al.(2022{\natexlab{b}})Montasser, Hanneke, and
  Srebro]{montasser_transductive_2022}
O.~Montasser, S.~Hanneke, and N.~Srebro.
\newblock Transductive {Robust} {Learning} {Guarantees}.
\newblock In \emph{Proceedings of {The} 25th {International} {Conference} on
  {Artificial} {Intelligence} and {Statistics}}, pages 11461--11471. PMLR, May
  2022{\natexlab{b}}.
\newblock URL \url{https://proceedings.mlr.press/v151/montasser22a.html}.
\newblock ISSN: 2640-3498.

\bibitem[Prost et~al.(2021)Prost, Houdard, Almansa, and
  Papadakis]{prost_learning_2021}
J.~Prost, A.~Houdard, A.~Almansa, and N.~Papadakis.
\newblock Learning {Local} {Regularization} for {Variational} {Image}
  {Restoration}.
\newblock In A.~Elmoataz, J.~Fadili, Y.~Quéau, J.~Rabin, and L.~Simon,
  editors, \emph{Scale {Space} and {Variational} {Methods} in {Computer}
  {Vision}}, Lecture {Notes} in {Computer} {Science}, pages 358--370, Cham,
  2021. Springer International Publishing.
\newblock ISBN 978-3-030-75549-2.
\newblock \doi{10.1007/978-3-030-75549-2_29}.

\bibitem[Rado(1967)]{rado_note_1967}
R.~Rado.
\newblock Note on the {Transfinite} {Case} of {Hall}'s {Theorem} on
  {Representatives}.
\newblock \emph{Journal of the London Mathematical Society}, s1-42\penalty0
  (1):\penalty0 321--324, 1967.
\newblock ISSN 00246107.
\newblock \doi{10.1112/jlms/s1-42.1.321}.
\newblock URL \url{http://doi.wiley.com/10.1112/jlms/s1-42.1.321}.

\bibitem[Rubinfeld et~al.(2011)Rubinfeld, Tamir, Vardi, and
  Xie]{rubinfeld_fast_2011}
R.~Rubinfeld, G.~Tamir, S.~Vardi, and N.~Xie.
\newblock Fast {Local} {Computation} {Algorithms}, Apr. 2011.
\newblock URL \url{http://arxiv.org/abs/1104.1377}.
\newblock arXiv:1104.1377 [cs].

\bibitem[Rubinstein et~al.(2009)Rubinstein, Bartlett, and
  Rubinstein]{rubinstein_shifting_2009}
B.~I.~P. Rubinstein, P.~L. Bartlett, and J.~H. Rubinstein.
\newblock Shifting: {One}-inclusion mistake bounds and sample compression.
\newblock \emph{Journal of Computer and System Sciences}, 75\penalty0
  (1):\penalty0 37--59, Jan. 2009.
\newblock ISSN 0022-0000.
\newblock \doi{10.1016/j.jcss.2008.07.005}.
\newblock URL
  \url{https://www.sciencedirect.com/science/article/pii/S0022000008000676}.

\bibitem[Schaffer(1994)]{schaffer_conservation_1994}
C.~Schaffer.
\newblock A {Conservation} {Law} for {Generalization} {Performance}.
\newblock In W.~W. Cohen and H.~Hirsh, editors, \emph{Machine {Learning}
  {Proceedings} 1994}, pages 259--265. Morgan Kaufmann, San Francisco (CA),
  Jan. 1994.
\newblock ISBN 978-1-55860-335-6.
\newblock \doi{10.1016/B978-1-55860-335-6.50039-8}.
\newblock URL
  \url{https://www.sciencedirect.com/science/article/pii/B9781558603356500398}.

\bibitem[Schrijver(2003)]{Schrijver_2003}
A.~Schrijver.
\newblock \emph{Combinatorial optimization: polyhedra and efficiency},
  volume~24.
\newblock Springer, 2003.
\newblock URL \url{https://link.springer.com/book/9783540443896}.

\bibitem[Shalev-Shwartz and Ben-David(2014)]{shalev-shwartz_understanding_2014}
S.~Shalev-Shwartz and S.~Ben-David.
\newblock \emph{Understanding {Machine} {Learning}: {From} {Theory} to
  {Algorithms}}.
\newblock Cambridge University Press, May 2014.
\newblock ISBN 978-1-139-95274-3.

\bibitem[Shalev-Shwartz et~al.(2010)Shalev-Shwartz, Shamir, Srebro, and
  Sridharan]{shalev-shwartz_learnability_2010}
S.~Shalev-Shwartz, O.~Shamir, N.~Srebro, and K.~Sridharan.
\newblock Learnability, stability and uniform convergence.
\newblock \emph{The Journal of Machine Learning Research}, 11:\penalty0
  2635--2670, 2010.

\bibitem[Tiegel(2024)]{tiegel_improved_2024}
S.~Tiegel.
\newblock Improved {Hardness} {Results} for {Learning} {Intersections} of
  {Halfspaces}.
\newblock In \emph{Proceedings of {Thirty} {Seventh} {Conference} on {Learning}
  {Theory}}, pages 4764--4786. PMLR, June 2024.
\newblock URL \url{https://proceedings.mlr.press/v247/tiegel24a.html}.
\newblock ISSN: 2640-3498.

\bibitem[Vapnik(1982)]{vapnik_estimation_1982}
V.~Vapnik.
\newblock \emph{Estimation of {Dependences} {Based} on {Empirical} {Data}}.
\newblock Springer {Series} in {Statistics}. Springer, New York, NY, 1982.

\bibitem[Vapnik and Chervonenkis(1974)]{vapnik_theory_1974}
V.~Vapnik and A.~Chervonenkis.
\newblock Theory of pattern recognition.
\newblock 1974.
\newblock Publisher: Nauka, Moscow.

\bibitem[Vapnik(1998)]{vapnik_nature_1998}
V.~N. Vapnik.
\newblock \emph{The {Nature} of {Statistical} {Learning} {Theory}}.
\newblock Springer, New York, NY, UNITED STATES, 1998.
\newblock ISBN 978-1-4757-2440-0.
\newblock URL
  \url{http://ebookcentral.proquest.com/lib/socal/detail.action?docID=3084784}.

\bibitem[Vaškevičius and Zhivotovskiy(2023)]{vaskevicius_suboptimality_2023}
T.~Vaškevičius and N.~Zhivotovskiy.
\newblock Suboptimality of constrained least squares and improvements via
  non-linear predictors.
\newblock \emph{Bernoulli}, 29\penalty0 (1):\penalty0 473--495, Feb. 2023.
\newblock ISSN 1350-7265.
\newblock \doi{10.3150/22-BEJ1465}.
\newblock URL
  \url{https://projecteuclid.org/journals/bernoulli/volume-29/issue-1/Suboptimality-of-constrained-least-squares-and-improvements-via-non-linear/10.3150/22-BEJ1465.full}.
\newblock Publisher: Bernoulli Society for Mathematical Statistics and
  Probability.

\bibitem[Wolf and Donner(2008)]{wolf_local_2008}
L.~Wolf and Y.~Donner.
\newblock Local {Regularization} for {Multiclass} {Classification} {Facing}
  {Significant} {Intraclass} {Variations}.
\newblock In D.~Forsyth, P.~Torr, and A.~Zisserman, editors, \emph{Computer
  {Vision} – {ECCV} 2008}, pages 748--759, Berlin, Heidelberg, 2008.
  Springer.
\newblock ISBN 978-3-540-88693-8.
\newblock \doi{10.1007/978-3-540-88693-8_55}.

\bibitem[Wolpert(1992)]{wolpert_connection_1992}
D.~H. Wolpert.
\newblock On the {Connection} between {In}-sample {Testing} and
  {Generalization} {Error}.
\newblock \emph{Complex Systems}, 6\penalty0 (1):\penalty0 47--94, 1992.

\bibitem[Wolpert(1996)]{wolpert_lack_1996}
D.~H. Wolpert.
\newblock The {Lack} of {A} {Priori} {Distinctions} {Between} {Learning}
  {Algorithms}.
\newblock \emph{Neural Computation}, 8\penalty0 (7):\penalty0 1341--1390, Oct.
  1996.
\newblock ISSN 0899-7667, 1530-888X.
\newblock \doi{10.1162/neco.1996.8.7.1341}.
\newblock URL \url{https://direct.mit.edu/neco/article/8/7/1341-1390/6016}.

\end{thebibliography}
\bibliographystyle{abbrvnat}               
}

\end{document}